\documentclass{article}

\usepackage{microtype}
\usepackage{graphicx}
\usepackage{booktabs} 
\usepackage{balance}

\usepackage[accepted]{icml2025}

\usepackage{amsmath}
\usepackage{amssymb}
\usepackage{mathtools}
\usepackage{amsthm}
\usepackage[acronym]{glossaries}
\newacronym{MEC}{MEC}{Minimum Entropy Coupling}

\newacronym{MAP}{map}{maximum-a-posteriori}
\newacronym{MLE}{mle}{maximum likelihood estimation}
\newacronym{MNLL}{mnll}{mean negative loglikelihood}
\newacronym{NLL}{nll}{negative loglikelihood}
\newacronym{LL}{ll}{log-likelihood}
\newacronym{RMSE}{rmse}{root mean square error}
\newacronym{ECE}{ece}{expected calibration error}
\newacronym{SNR}{snr}{signal-to-noise ratio}
\newacronym{FID}{\textsc{fid}}{Fr\'echet Inception Distance}
\newacronym{SSIM}{\textsc{ssim}}{structural Similarity Index Measure}
\newacronym{FAD}{fad}{Fr\'echet Audio Distance}
\newacronym{FMD}{fmd}{Fr\'echet Modality Distance}
\newacronym{BPD}{bpd}{bit per dimension}
\newacronym{NFE}{nfe}{neural function evaluations}

\newacronym{AE}{ae}{autoencoder}
\newacronym{WAE}{wae}{Wasserstein Autoencoder}
\newacronym{VAE}{vae}{Variational Autoencoder}
\newacronym{BAE}{bae}{Bayesian autoencoder}
\newacronym{CDF}{cdf}{cumulative density function}
\newacronym{GAN}{\textsc{gan}}{Generative Adversarial Network}
\newacronym{DPGMM}{dpgmm}{Dirichlet process Gaussian mixture model}

\newacronym{MC}{mc}{Monte Carlo}
\newacronym{SDE}{sde}{Stochastic Differential Equation}
\newacronym{CNF}{cnf}{Continuous Normaxlizing Flow}
\newacronym{ODE}{ode}{Ordinary Differential Equation}
\newacronym{MCMC}{mcmc}{Markov chain Monte Carlo}
\newacronym{HMC}{hmc}{Hamiltonian Monte Carlo}
\newacronym{MH}{mh}{Metropolis-Hastings}
\newacronym{NUTS}{nuts}{no-u-turn sampler}
\newacronym{SGHMC}{sghmc}{stochastic gradient Hamiltonian Monte Carlo}

\newacronym{DGP}{dgp}{deep Gaussian process} 
\newacronym{GPLVM}{gplvm}{Gaussian process latent variable model}
\newacronym{DPMM}{dpmm}{Dirichlet Process Mixture Model}

\newacronym{VFE}{vfe}{variational free energy}

\newacronym[firstplural=Gaussian Processes]{GP}{gp}{Gaussian Process}

\newacronym{VI}{vi}{variational inference}

\newacronym{PDE}{pde}{Partial Differential Equation}
\newacronym{ELBO}{elbo}{evidence lower bound}
\newacronym{NELBO}{nelbo}{negative evidence lower bound}
\newacronym{ELL}{ell}{expected log likelihood}
\newacronym{KL}{kl}{Kullback-Leibler}
\newacronym{AUC}{auc}{area under the curve}

\newacronym[firstplural=Bayesian neural networks]{BNN}{bnn}{Bayesian neural network}
\newacronym[firstplural=deep neural networks]{DNN}{dnn}{deep neural network}
\newacronym[]{CNN}{cnn}{convolutional neural network}
\newacronym{MLP}{mlp}{multilayer perceptron}
\newacronym{NN}{nn}{neural network}
\newacronym{RELU}{ReLU}{rectified linear unit}

\newacronym{NF}{nf}{normalizing flow}

\newacronym{RBF}{rbf}{radial basis function}
\newacronym{ARD}{ard}{automatic relevance determination}

\newacronym{RKHS}{rkhs}{reproducing kernel Hilbert space}

\newacronym{OT}{OT}{Optimal Transport}
\newacronym{WD}{wd}{Wasserstein distance}
\newacronym{SWD}{swd}{sliced-Wasserstein distance}
\newacronym{DSWD}{dswd}{distributional sliced-Wasserstein distance}
\newacronym{RL}{RL}{Reinforcement Learning}

\newacronym{MLD}{mld}{Multi-modal Latent Diffusion}
\newacronym{MLD Inpaint}{mld in-paint}{Multi-modal Latent Diffusion with In-painting}
\newacronym{MLD Uni}{mld uni}{Multi-modal Latent Diffusion UniDiffuser }
\newacronym{MOPOE}{mopoe}{Mixture of Product of Experts}
\newacronym{MVAE}{mvae}{Product of Experts}
\newacronym{MMVAE}{mmvae}{Mixture of Expert}
\newacronym{NEXUS}{nexus}{Hierarchical Genertive Model}

\newacronym{MMVAEplus}{MMVAE+}{}

\newacronym{MVTCAE}{mvtcae}{Multi-view Total Correlation Autoencoder}
\newacronym{CLIP-S}{clip-s}{CLIP-Score}
\newacronym{MHD}{mhd}{The Multimodal Handwritten Digits data-set}
\newacronym{VPSDE}{vpsde}{Variance preserving SDE}
\newacronym{EMA}{ema}{Exponential moving average}
\newacronym{MSE}{mse}{Mean square error}
\newacronym{MED}{MED}{Minimum Entropy Diffusion}
\newacronym{DDPM}{DDPM}{Denoising Diffusion Probabilistic Models }
\newacronym{GANS}{GANS}{Generative Adversarial Networks }
\newacronym{PCA}{PCA}{Principal Component Analysis
 }
\newcommand{\ddmec}{\textsc{DDMEC }}

\DeclareRobustCommand{\KL}[2]{{\mathbb{KL}\left[#1\;\|\;#2\right]}}
\DeclarePairedDelimiterX{\infdivx}[2]{[}{]}{%
#1\;\delimsize\|\;#2%
}

\DeclareMathOperator*{\argmin}{arg\,min}

\newcommand{\cN}{\mathcal{N}}
\newcommand{\cP}{\mathcal{P}}

\newcommand{\cX}{\mathcal{X}}
\newcommand{\cY}{\mathcal{Y}}

\newcommand{\cU}{\mathcal{U}}

\newcommand{\E}{\mathbb{E}}

\usepackage{hyperref}
\usepackage[capitalize,noabbrev]{cleveref}

\usepackage{flushend}

\theoremstyle{plain}
\newtheorem{theorem}{Theorem}[section]

\theoremstyle{definition}
\newtheorem{definition}[theorem]{Definition}

\theoremstyle{remark}

\usepackage[disable,textsize=tiny]{todonotes}
\usepackage{subcaption}

\icmltitlerunning{Learning to Match Unpaired Data with Minimum Entropy Coupling}

\begin{document}

\twocolumn[
\icmltitle{Learning to Match Unpaired Data with Minimum Entropy Coupling}




\begin{icmlauthorlist}
\icmlauthor{Mustapha Bounoua}{comp,un}
\icmlauthor{Giulio Franzese}{un}
\icmlauthor{Pietro Michiardi}{un}
\end{icmlauthorlist}

\icmlaffiliation{comp}{Ampere Software Technology, France}
\icmlaffiliation{un}{Department of Data Science, EURECOM, France}

\icmlcorrespondingauthor{}{mustapha.bounoua@eurecom.fr}

\icmlkeywords{Machine Learning, ICML, Diffusion models, Multimodality, Unpaired data, Minimum entropy coupling}

\vskip 0.3in
]



\printAffiliationsAndNotice{}  

\begin{abstract}
Multimodal data is a precious asset enabling a variety of downstream tasks in machine learning. However, real-world data collected across different modalities is often not paired, which is a significant challenge to learn a joint distribution. A prominent approach to address the modality coupling problem is Minimum Entropy Coupling (MEC), which seeks to minimize the joint Entropy, while satisfying constraints on the marginals. Existing approaches to the MEC problem focus on finite, discrete distributions, limiting their application for cases involving continuous data. In this work, we propose a novel method to solve the continuous MEC problem, using well-known generative diffusion models that learn to approximate and minimize the joint Entropy through a cooperative scheme, while satisfying a relaxed version of the marginal constraints.
We empirically demonstrate that our method, \ddmec, is general and can be easily used to address challenging tasks, including unsupervised single-cell multi-omics data alignment and unpaired image translation, outperforming specialized methods.

\end{abstract}

\section{Introduction}\label{sec:introduction}

Nowadays, multimodal data is pervasive thanks to advances in data collection technologies and the crucial need for systems that can learn from the diversity of real-world phenomena. Healthcare, for example, is a domain where patient data often spans electronic health records, radiological images, genetic data, and wearable sensor outputs~\citep{kline2022multimodal, acosta2022multimodal}. Autonomous systems rely on a suite of sensors, including LiDAR, cameras, and ultrasonic sensors, to navigate environments effectively~\citep{caesar2020nuscenes, gu2023end, franchi2024infraparis}. Scientific disciplines, such as astronomy and geoscience, employ multimodal datasets combining spatial, spectral, and temporal data to understand complex systems~\citep{srivastava2019understanding, zhang2024when, sosic2024multiscale}.

Modeling multimodal data allows for a more comprehensive understanding, reflecting the inherently multi-faceted nature of the real world. Recent works in representation learning~\citep{radford2021learning, zhou2023theory, manzoor2023multimodality, chen2023enhanced}, the study of multivariate systems~\citep{kaplanis2023learning, liang2023quantifying, bounoua2024somegai}, generative modeling~\citep{rombach2022high, tang2023any, tang2023codi2, bounoua2024multi, esser2024scaling}, and multimodal conversational agents~\citep{li2023blip2, liu2023visual, shukor2023unival, xue2024xgenmmblip3familyopen, wu2024nextgpt}, are few examples to illustrate the fervent effort in the machine learning community to address and exploit multimodality.
However, the intrinsic complexity of multimodal data introduces several challenges that hinder their application in machine learning research. Modality heterogeneity complicates and sometimes impedes geometric comparisons, requiring for example learning a mapping from ambient to latent spaces~\citep{rombach2022high, tang2023any, liu2023visual, bounoua2024multi} or stringent assumptions~\citep{liang2022mind, xia2023achieving, dong2024dreamllm, ibrahimi2024intriguing, zhang2024connect}. Alignment across modalities at spatial, temporal or semantic levels is another challenge, which calls for costly pre-processing steps such as synchronization~\citep{hanchate2024process, chen2024terra, scire2024emergence, turrero2024alert}. 

The major roadblock we address is that of paired multimodal data, which is an underlying assumption in many works in the literature~\citep{radford2021learning, rombach2022high, liu2023visual, li2023blip2, bounoua2024multi}. Paired data -- for a given sample, all its various modalities are available -- is either expensive, difficult to obtain, or sometimes impossible. 
For example, in genetic research, data is inherently unpaired due to the nature of the data acquisition process, such as single-cell RNA sequencing data, where measurements destroy the original cells~\citep{kester2018single, chen2019high, schiebinger2019optimal}. 
Similarly, matching image data from different domains is a challenging endeavor when paired data is missing, which calls for specialized methods~\cite{zhu2017unpaired, huang2018multimodal, pang2021image, sasaki2021unit, yang2023gp, sun2023sddm, Xie_2023_CVPR}.

In this work, we study the problem of unpaired multimodal data through the lens of \textit{coupling}, a fundamental problem in probability theory, that aims at \textit{determining the optimal joint distribution of random variables given their marginal distributions}, with early attempts at solving it dating back to the work by~\citet{frechet1951sur}. 
The pairing problem belongs to a broad class of methods~\citep{denhollander2012coupling, lin2014recent, benes2012distributions, yu2018asymptotic}: some cast it through the lens of information-theoretic quantities, where optimality is defined in terms of Entropy minimization or Mutual Information maximization, others focus on \gls{OT}~\citep{villani2009optimal, peyre2019computational}, where optimality is defined as minimizing the expected value of a transport cost over the joint distribution.
Our focus is the \gls{MEC} problem, which aims at finding the joint distribution with the smallest Entropy, given the marginal distribution of some random variables. 
Recent applications include entropic causal inference~\citep{kocaoglu2017entropic, javidian2021quantum, compton2022tighter}, communication systems~\citep{sokota2022communicating}, steganography~\citep{de2022perfectly}, random number generation~\citep{li2021efficient}, dimensionality reduction~\citep{cicalese2016approximating, vidyasagar2012metric}, lossy compression~\citep{ebrahimi2024minimum}, and multimodal learning~\citep{liang2024multimodal}. 

While the \gls{MEC} problem is known to be NP-Hard~\citep{vidyasagar2012metric, kovavcevic2012hardness}, the literature contains many approximation and greedy algorithms 
~\citep{painsky2013memoryless, Kovacevic, cicalese2016approximating, li2021efficient}, and theoretical studies about the approximation qualities of such approaches~\citep{cicalese2017joint, cicalese2019minimum}. Nevertheless, the vast majority of prior work on the \gls{MEC} problem focus on discrete distributions: instead, we consider the continuous variant of \gls{MEC}, and propose a flexible and general solution to the coupling problem for arbitrary, continuous distributions. The \gls{MEC} problem for continuous random variables is much more complex than its discrete counterpart, and can be ill-defined in certain cases due to the properties of differential Entropy and the challenges inherent to continuous distributions living in an infinite dimensional space.

The gist of our method is to consider a parametric class of joint distributions, which we reinterpret as conditional generative models, with additional terms to steer adherence to marginal constraints. Then, the \gls{MEC} problem requires access to the conditional Entropy, which we rewrite as log-likelihood. Crucially, our method exploits two specular generative models, which cooperate to minimize the joint Entropy, while approximately satisfying the marginal constraints. Our approach materializes as two denoising diffusion probabilistic models~\cite{ho2020denoising}, which we first pre-train on marginal distributions, and then fine-tune according to reward functions, following an alternating optimization process.
In summary, our contributions are:
\begin{itemize}
    \item We propose an approximation of the \gls{MEC} problem for arbitrary, continuous distributions, which is general, and that does not require stringent assumptions on the marginal distributions, nor the definition of geometric cost functions (\Cref{sec:problem-formulation}).
    \item We present a practical implementation of our method (\Cref{sec:method}), that relies on generative models, that interact through a cooperative scheme aiming at optimizing an information-theoretic cost function related to the Entropy of the joint distribution. Our training procedure overcomes numerical instabilities and degenerate solutions by relying on the application of soft marginal constraints, as well as the natural approximation stemming from a finite-capacity denoising model.
    \item We illustrate the benefits and performance of our method on two important use cases (\Cref{sec:experiments}). First, we solve the coupling problem between incomparable spaces with a single-cell multi-omics dataset, where we compare our method to state-of-the-art alternatives that rely on \gls{OT}. Second, we focus on unsupervised image translation between uncoupled pairs, and compare against state of the art. 
\end{itemize}

\section{Problem Formulation}\label{sec:problem-formulation}
Given two random variables $X \in \cX$ and $Y \in \cY$ with marginal probability distributions $p_X(x)$ and $p_Y(y)$ respectively, we consider a \textit{parametric} space $\cP^{\theta}=\{p^{\theta}_{XY}(x, y)\}$ of \textit{joint} distributions over the space $\cX \times \cY$, with induced marginal distributions $p^{\theta}_X(x),p^{\theta}_Y(y)$ (where $p^{\theta}_X(x) \triangleq \int_{\cY} p^{\theta}_{XY}(x, y) \, dy $ and similarly for $p^{\theta}_Y(y)$). 
The \gls{MEC} problem between the two original distributions $p_X(x)$ and $p_Y(y)$ consists in finding a joint distribution $p^{\theta}_{XY}(x, y)$ such that i) the induced marginal distributions $p^{\theta}_X(x),p^{\theta}_Y(y)$ match them either exactly or approximately and ii) the joint distribution is the one with minimal entropy~\cite{Kovacevic, cicalese2017joint, cicalese2019minimum}. The constraints over the search space $\cP^{\theta}$ are referred to as \textit{marginal contraints} 
\begin{definition}
    A joint distribution $p^{\theta}_{XY}(x, y)$ from $\cP^\theta$ is said to be an \textit{exact} coupling iff 
    \begin{equation}
        p^{\theta}_X(x)=p_X(x),p^{\theta}_Y(y)=p_Y(y).
    \end{equation}
\end{definition}
In general, exact coupling is not possible (nor wanted, to avoid overfitting) and the goodness of the solution in terms of marginal constraints is approximated through some distance function between the induced and original distributions, e.g. using the Kullback-Leibler divergence $\KL{p^{\theta}_X}{p_X}\triangleq \E_{x\sim p^{\theta}_X}\left[\log\frac{p^{\theta}_X}{p_X}(x)\right]$. Then, we define the \gls{MEC} problem with \textit{soft} constraints as follows 
\begin{equation}\label{def:coupling-generic} 
         \min_{\theta} \mathbb{H}(p^{\theta}_{XY})+\lambda_X \KL{p^{\theta}_X}{p_X}+\lambda_Y \KL{p^{\theta}_Y}{p_Y},
 \end{equation}

where the entropic term is defined as 
$
\mathbb{H}(p^{\theta}_{XY}) \triangleq -\E_{x,y\sim p^{\theta}_{X,Y}}\left[ \log p^\theta_{XY}(x, y)\right]
$. 

Previous work have mainly focused on the exact \gls{MEC} in discrete settings, where $p_X$ and $p_Y$ have a finite or countably infinite number of outcomes. Exact solution in such settings is known to be NP-Hard~\citep{Vidyasagar, Kovacevic}. Under our assumption of continuous spaces the problem is more complex. Exact matching is not generally possible due to the finite complexity of the parametric family $\cP^\theta$, since in general the distributions $p_X,p_Y$ live in infinite dimensional spaces. Rather than a limitation, enforcing limited complexity is helpful to avoid degenerate, deterministic joint probabilities (e.g. $p^{\theta}_{XY}(x,y)=\delta(y-g(x))p_{X}(x)$, where $g(\cdot)$ is any mapping which guarantees exact coupling), which would induce infinite joint entropy.

Interestingly, the \gls{MEC} problem has an intuitive interpretation connected to the problem of Mutual Information maximization. Indeed, $\mathbb{I}(p^{\theta}_{XY}) \triangleq -\mathbb{H}(p^{\theta}_{XY}) + \mathbb{H}(p^{\theta}_{X}) + \mathbb{H}(p^{\theta}_{Y})$ and in the exact matching scenario $\mathbb{H}(p^{\theta}_{X})=\mathbb{H}(p_X), \mathbb{H}(p^{\theta}_{Y})=\mathbb{H}(p_Y)$.  In other words, whenever the marginal constraints are satisfied with reasonable quality, the \gls{MEC} problem is a good approximation of the information maximization problem.

Early instances of the coupling problem express it through the lenses of \gls{OT}~\citep{monge1781memoire, kantorovich1942translocation}. In the simplest (albeit rich and interesting) scenario, the goal is to minimize the transportation cost between distributions $\E_{x,y\sim p^{\theta}_{X,Y}}[||x-y||^2]$, with the implicit assumption of $\cX=\cY=\mathbb{R}^N$ and under the requirement of exact matching (corresponding to $\lambda_X=\lambda_Y=\infty$). Several interesting extensions, including additional constraints on the joint distribution such as geometry or structural constraints, lead to tailor-made approaches~\citep{villani2009optimal, peyre2019computational}. Other than the trivial relaxation of constraints from exact to approximate, a particularly useful extension concerns the \textit{entropy-regularized} version of this problem, where the cost function is complemented by the entropic term $
\mathbb{H}(p^{\theta}_{XY})$. Although \gls{MEC} is fundamentally different than \gls{OT}, a link between the two clearly exists. However, a straightforward comparison is not possible, as the entropic term enters the respective minimization problems with different signs. Minimizing $\mathbb{H}(p^{\theta}_{XY})$ directly over other geometric costs (like the euclidean norm considered in \gls{OT}) has several advantages in terms of generality, as it does not require geometrically comparable spaces $\cX$ and $\cY$.

\section{Methodology}\label{sec:method}

Consider two random variables in continuous domains, $X \in \cX$ and $Y \in \cY$. We begin by considering a parametric class for the joint distribution expressed as $p_{X,Y}^{\theta}(x,y)= p^{\theta}_{X| Y}(x|y)p_Y(y)$, such that the joint entropy $\mathbb{H}(p_{X,Y}^{\theta})$ minimization becomes equivalent to minimizing the conditional entropy $\mathbb{H}(p^{\theta}_{X| Y=y})$. Note that the marginal constraint on $Y$ from \Cref{def:coupling-generic} is verified by construction. To satisfy the marginal constraint on $X$ we consider the $\mathbb{KL}$ divergence. 
This leads to an alternative definition of the \gls{MEC} problem with soft constraints, that reads as 

\begin{definition}\label{def:MEC-approx}
    Given random variables $X \in \cX$ and $Y \in \cY$, the continuous \gls{MEC} problem with soft marginal constraints corresponds to the optimization problem
    \begin{equation}\label{eq:MEC-optim}
        \min_{\theta} \E_{y\sim p_Y}\left[\mathbb{H}(p^\theta_{X| Y=y})\right] + \lambda_X \KL{p^{\theta}_X}{p_X}.
    \end{equation}
\end{definition}

Crucially, we note that the parametric portion of the joint distribution, namely $p^{\theta}_{X| Y}(x|y)$, can be interpreted as a conditional generative model of the variable $X$ given $Y$. As a consequence, the conditional entropy from \Cref{def:MEC-approx}, can be interpreted as an expected log-likelihood, leading to 

\begin{equation}\label{eq:MEC-llik-optim}
    \min_{\theta} 
    \E_{\substack{x,y \sim p^\theta_{X,Y}}}
    \left[ - \log \left(p^\theta_{X| Y=y}\right)\right] 
    + \lambda_X \KL{p^{\theta}_X}{p_X}.
\end{equation}

A maximum likelihood solution to the \gls{MEC} problem in \Cref{eq:MEC-llik-optim} is appealing, because it can be addressed by learning the parameters of an appropriate \textit{conditional} generative model, while approximating the marginal constraints on $X$ through the unconditional version of the model. Nevertheless, this approach bears several challenges:
\begin{itemize}
    \item Asymmetry: \Cref{eq:MEC-llik-optim} can be used to minimize the conditional entropy $\mathbb{H}(p^\theta_{X| Y=y})$. The learned conditional generative model can be used to generate samples from variable $X$ given $Y$, but not vice-versa. 
    \item Marginal constraint: in principle, exact matching requires $\lambda_X \to \infty$, but this choice leads to degenerate solutions to the \gls{MEC} problem. The marginal constraint from \Cref{def:MEC-approx}, despite being \textit{soft}, should strive to keep $p^{\theta}_X$ anchored to $p_X(x)$, which is not known. 
\end{itemize}

To address the first challenge, we introduce a second family of parametric models $p^{\phi}_{Y| X}(y|x)p_X(x)$, this time corresponding to conditional generative model of the variable $Y$ given observations of $X$. Then, we can write a specular version of the \gls{MEC} problem we defined as 

\begin{equation}\label{eq:MEC-llik-optim-specular}
    \min_{\phi} 
    \E_{\substack{x,y\sim p^\phi_{X,Y}}}
    \left[ - \log \left(p^\phi_{Y| X=x}\right)\right]  
    + \lambda_Y \KL{p^{\phi}_Y}{p_Y}.
\end{equation}

Recall that $p^{\theta}_{X,Y}=p^{\theta}_{X|Y}p_Y$ and $p^{\phi}_{X,Y}=p^{\phi}_{Y|X}p_X$: it is then reasonable to strive, among all the possible solutions, for $p^{\theta}_{X,Y} = p^{\phi}_{X,Y}$. This \textit{joint constraint} can be approximated with a penalty term proportional to the $\mathbb{KL}$ divergence between the two distributions. Interestingly, this coupling allows to implement a practical method that exploits cooperation: we use $p^{\phi}_{Y|X}$ to improve $p^{\theta}_{X|Y}$, and vice-versa.

To address the second challenge, and pave the way for our practical implementation, we break the optimization problem by first focusing on respecting the marginal constraints. To do so, we pretrain unconditional models such that $p^{\theta_{*}}_X(x) \approx p_X(x)$ and $p^{\phi_{*}}_Y(y) \approx p_Y(y)$. Then, we use $\theta_{*}$ and $\phi_{*}$ to initialize the conditional models, and anchor their parameters throughout the optimization such that they do not deviate too much from the pretrained models.

Overall, the method we propose writes as

\begin{align}\label{eq:approx-MEC-abstract} 
     {} &\min\limits_{\theta} 
     \E_{\substack{x,y\sim  p^\theta_{X,Y}}}
     \left[ 
        - \log \left(p^\phi_{Y| X=x}\right)
    \right]
    +\lambda_X \KL{p^{\theta}_{X}}{p^{\theta_{*}}_X}, \nonumber\\
     {}& \min\limits_{\phi} 
     \E_{\substack{x,y\sim p^\phi_{X,Y}}}
     \left[ 
        - \log \left (p^\theta_{X| Y=y}\right)
    \right]
    +\lambda_Y \KL{p^{\phi}_{Y}}{p^{\phi_{*}}_Y},
\end{align}

where we additionally enforce the approximate joint constraint. 
Notice the difference with \Cref{eq:MEC-llik-optim} and \Cref{eq:MEC-llik-optim-specular}: given the structure of the parametric distributions we use, it is possible to show that $\nabla_\theta \E_{x,y\sim p^\theta_{X,Y}}
[ - \log p^{\theta}_{X|Y}] \approx \nabla_\theta \E_{x,y\sim p^\theta_{X,Y}} [- \log p^{\phi}_{Y|X}]$ whenever $p^\theta_{X,Y} = p^\phi_{X,Y}$ (see \Cref{apdx:proof} for more details). Strict adherence to the joint constraint, in principle, allows ``swapping'' the roles of the conditional models without affecting the optimization dynamics, leading to a cooperative method. In practice, we found trough empirical exploration that such a cooperative formulation (See \Cref{ddmec_scheme}), albeit approximate, proves to be much more stable than the original problem, and consequently decided to adopt it in our implementation, as described next.

\subsection{Practical implementation}\label{sec:diffusion-models}
In our implementation, we consider the parametric class of probability distributions associated to denoising diffusion probabilistic models (DDPM)~\citep{Sohldiff,ho2020denoising}. These models enjoy excellent performance in fitting complex multimodal data, and allow accurate estimation of information metrics~\cite{minde,kong2023interpretable,bounoua2024somegai,dewan2024diffusion}. 

\paragraph{DDPM.} These generative models are characterized by a forward process, that is fixed to a Markov chain that gradually adds Gaussian noise to the data according to a carefully selected variance schedule $\beta_t$, i.e. $x_t=\sqrt{1-\beta_t}x_{t-1}+\sqrt{\beta_t \epsilon}$ with $\epsilon\sim \cN(0,I)$. Interestingly, an arbitrary portion of this forward chain can be efficiently simulated through the equality in distribution $x_t=\sqrt{\bar{\alpha}_t} x_0 + (\sqrt{1-\bar{\alpha}_t}) \epsilon$, with $x_0\sim p_X$ and $\alpha_t = 1 - \beta_t$, $\bar{\alpha}_t = \prod_{s=1}^{t} \alpha_s$.

The corresponding discrete-time reverse process, that has a Markov structure as well, is used for generative purposes. The model generates data through the iterative sampling process $p_{X}^\theta(x_{0\dots T})=\prod\limits_{t=1}^{T}  p^\theta(x_{t-1} | x_t) p^\theta(x_T)$,
where $p^\theta(x_T)=\mathcal{N}(x_T; 0, I)$ and typically $p^{\theta}({x}_{t-1} | {x}_t)$ is a Gaussian transition kernel with mean $\frac{1}{\sqrt{\alpha_t}} \left( {x}_t - \frac{\beta_t}{\sqrt{1 - \bar{\alpha}_t}} \epsilon^{\theta}({x}_t, t) \right)$ and covariance $\beta_t I$. Intuitively, starting from a simple distribution $x_T \sim \cN(\mathbf{0}, \mathbf{I})$, samples are generated by a denoising network $\epsilon^\theta$, that removes noise over $T$ denoising steps.
A simple way to learn the denoising network $\epsilon^\theta$ is to consider a re-weighted variational lower bound of the expected marginal likelihood, where the problem $\argmin_{\theta} \KL{p_X}{p^\theta_X}$ becomes

\begin{align}\label{eq:lsimple}
    &  \argmin_{\theta}\sum_{t=1}^T \E_{\epsilon\sim\cN(0,I),x_0\sim p_X} \left[ || \epsilon - \epsilon^\theta (x_t, t) ||^2 \right].
\end{align}

This simple formulation has been extended to conditional generation~\citep{ho2021classifierfree}, whereby a conditioning signal $y$ injects ``external information'' in the iterative denoising process. This requires a simple extension to the denoising network such that it can accept the conditioning signal: $\epsilon^\theta(x_t, y, t)$. During training, a randomized approach allows to learn both the conditional and unconditional variants of the denoising network, for example by assigning a null value to the conditioning signal, e.g. $y=\emptyset$.

In \Cref{eq:approx-MEC-abstract}, the log-likelihood emerges as a critical quantity to address the \gls{MEC} problem. In the ideal conditions of a \textit{perfect} denoising network, the difference between predicted and actual noise can be used, in the limit of infinite number of denoising steps, to compute exactly such quantity~\cite{kong2023informationtheoretic}. We use these results to compute the log-likelihoods through Monte Carlo estimation techniques 

\begin{align}
\label{eq:likelihood}
-\log p_\theta(x_0) \approx \text{const} + \frac{1}{2} \sum_{t=1}^T \E_{\epsilon} \left[\|\epsilon - \epsilon^\theta(x_t, t)\|_2^2\right],
\end{align}
where the unspecified constant does not depend neither on $x_0$ nor on $\theta$, and is consequently irrelevant for optimization purposes. This approach can be trivially generalized to the case of a conditional denoising network $\epsilon^\theta(x_t, y, t)$.

\paragraph{Our method: \ddmec.} We being by pretraining \textit{unconditional} models such that $p^{\theta_{*}}_X(x) \approx p_X(x)$ and $p^{\phi_{*}}_Y(y) \approx p_Y(y)$. Then, we use $\theta_{*}$ and $\phi_{*}$ to initialize conditional models $p^{\theta}_{X|Y}$ and $p^{\phi}_{Y|X}$, which use denoising networks that accept additional conditioning signals, following~\citet{zhang2023adding}.
Next, we interpret the optimization expressed in \Cref{eq:approx-MEC-abstract} as a model fine-tuning objective, which is reminiscent of the work by~\citet{dpok}.

\begin{align}\label{eq:DDMEC} 
    {}&
    \min\limits_{\theta} 
    \E_{x,y\sim p^\theta_{X,Y}}
       r^{\phi}(y,x)    
    +\tilde{\lambda}_X \KL{p^{\theta}_{X}}{p^{\theta_{*}}_X}, \nonumber
    \\
    {}&
    \min\limits_{\phi} 
    \E_{x,y\sim p^\phi_{X,Y}}
       r^{\theta}(x,y)
    +\tilde{\lambda}_Y \KL{p^{\phi}_{Y}}{p^{\phi_{*}}_Y},
\end{align}

where \( r^{\phi} = - \log p^{\phi}_{Y|X} \) and \( r^{\theta} = - \log p^{\theta}_{X|Y} \) are reward signals striving to minimize the conditional entropies, and $\tilde{\lambda}_X, \tilde{\lambda}_Y$ are scaling factors used for fine-tuning, that no longer require to be extremely large. Furthermore, we enforce the joint constraints via extra penalty terms $\KL{p^\theta_{X,Y}}{p^\phi_{X,Y}},\KL{p^\phi_{X,Y}}{p^\theta_{X,Y}}$. 

Fine-tuning DDPMs introduces significant computational overhead. To address this, various studies have explored supervised methods~\citep{lee2023aligning, Wu_2023_ICCV} or reinforcement learning. In~\citep{clark2023directly, xu2024imagereward}, fine-tuning is achieved through direct back-propagation through the reverse process, which can be costly. Alternative methods use proximal policy optimization (PPO)~\cite{dpok, dppo, uehara2024fine}, leading to improved stability. Note that
\cite{dpok} incorporates KL-regularization to maximize the reward signal, while ensuring fidelity to the pretrained model, which is analogous to our soft marginal constraints.

In our implementation, we compute gradients of the reward $\nabla_{\theta} 
    \E_{ p^\theta_{X,Y}}
        r^{\phi}(y,x)$ as follows~\cite{dpok}
\begin{align}\label{eq:reward-grad}
    \E_{p^\theta_{X,Y}}
        r^{\phi}(y,x)
        \sum_{t=1}^{T}
        \nabla_{\theta}
        \log p^{\theta}(x_{t-1}|x_t,y)
\end{align}

while the gradient of the marginal constraints $\nabla_{\theta}
\KL{p^{\theta}_{X}}{p^{\theta_{*}}_X}$ are obtained as the approximate gradient of an upper bound~\cite{dpok}

\begin{align}\label{eq:KL-grad}
&
\sum_{t=1}^{T} 
\nabla_{\theta} 
\E_{x_t}\left[ ||\epsilon^\theta(x_t,y,t)-\epsilon^{\theta_*}(x_t,t) ||^2\right]
\end{align}
Similar expressions apply to the specular model.

\begin{algorithm}[tb]
\caption{\ddmec~ Training Loop}\label{algo:trainloop}
\begin{algorithmic}
\STATE \textbf{Input:} \( \theta_*,\phi_* \)
\STATE Initialize \( \theta\leftarrow\theta_*,\phi\leftarrow\phi_* \)
\REPEAT
\STATE Call \Cref{algo:trainstep} with \( y\sim p_Y, \theta, \theta_{*},\phi\)
\STATE Call \Cref{algo:trainstep} with \(x\sim p_X, \phi,\phi_{*},\theta\)
\UNTIL{Converged}
\end{algorithmic}
\end{algorithm}

Given pretrained models $p^{\theta_{*}}_X, p^{\phi_{*}}_Y$, the pseudo-code of our \ddmec~method in~\Cref{algo:trainloop} is extremely simple, as it materializes as alternating optimization steps, described (for the top~\Cref{eq:DDMEC}) in~\Cref{algo:trainstep}. 
First, we optimize for the parameters $\theta$ of the model $p^\theta_{X| Y=y}$, while fixing the parameters $\phi$ of the specular model $p^\phi_{Y|X}$, which we use as a \textit{reward} term. Then we adapt the parameters $\phi$ to ensure $p^{\phi}_{X,Y} \approx p^{\theta}_{X,Y}$: this is achieved by noting that we can adapt \Cref{eq:lsimple} to this purpose, whereby the parameters $\theta$ are now fixed.
In the second phase (which can described as the specular version of~\Cref{algo:trainstep}), we optimize for the parameters $\phi$ of the model $p^\phi_{Y|X=x}$, while fixing the parameters $\theta$ of the model $p^\theta_{X| Y}$ using the corresponding reward term. Finally, in a specular manner to the first phase, we ensure coherency of the two models by adapting $\theta$ such that $p^{\theta}_{X,Y} \approx p^{\phi}_{X,Y}$, thus satisfying the joint constraint. 

\begin{algorithm}[tb]
\caption{\ddmec~ Training Step}\label{algo:trainstep}
\begin{algorithmic}
\STATE \textbf{Input:} \( y, \theta, \theta_{*},\phi  \)
\STATE \( x \sim p^{\theta}_{X|Y=y}, t\sim \cU[0,T],\epsilon\sim\cN(0,I)\)\label{line2} 
\STATE Update $\theta$ using \Cref{eq:reward-grad,eq:KL-grad} \label{line3} 
\STATE Update $\phi$ using
\(
\nabla_{\phi}\mathbb{E}_{y_t, t} \big[ \| {\epsilon} - {\epsilon}_{\phi}(y_t, x, t) \|^2 \big]
\) \label{line4}
\end{algorithmic}
\end{algorithm}

\section{Experiments}\label{sec:experiments}
\ddmec~is a general method that can be applied across a variety of data domains, as it relies on an information-theoretic measure to match unpaired entities. Next, we demonstrate \ddmec~versatility using two realistic pairing tasks that use various data modalities, including multi-omics and image data. We compare \ddmec~to state-of-the-art methods for each task, and measure performance using domain-specific metrics.
More details about \ddmec~implementation, and our experimental protocol are given in \Cref{app:implementation_details}.

\subsection{Multi-omics single-cell alignment}

Single-cell measurements techniques, such as mRNA sequencing for whole-transcriptome analysis at the single-cell level~\cite{tang2009mrna}, have been adapted and commercialized by companies which developed platforms to facilitate scalable and efficient single-cell transcriptomics and multi-omics data collection. This data provides a detailed snapshot of the heterogeneous landscape of cells in a sample, and can be used to study the cell developmental trajectories across time, for example. The availability of multi-omics measurements -- capturing various properties of a cell, such as gene expression, mRNA transcriptomes, chromatin accessibility, histone modifications, to name a few -- calls for data integration methods to combine a variety of modalities~\citep{xi2024propensity}. Unfortunately, current measurement techniques are destructive: it is hard to obtain multiple types of measurements from the same cell. Furthermore, it is well-known that different cell properties, such as transcriptional and chromatin profiles, cannot be matched using the geometric properties of features in the two domains. Then, pairing single-cell data modalities requires methods that do not rely on either common cells or common features across the data types~\cite{welch2017matcher, amodio2018magan, welch2019single, stuart2019comprehensive}.

\paragraph{Baselines.} We compare our proposed method \ddmec~to several baselines from both the machine learning and bioinformatics literature, including 
\textsc{scot}~\cite{demetci2022scot}, 
\textsc{mmd-ma}~\cite{liu2019jointly}, 
\textsc{unioncom}~\cite{cao2020unsupervised}, 
and \textsc{scTopoGAN}~\cite{singh2023sctopogan}. 
\textsc{scot} proposes a variant of an \gls{OT} formulation based on the Gromov-Wasserstein distance, which preserves local neighborhood geometry when transporting data points. 
\textsc{mmd-ma} is a global manifold alignment algorithm based on the maximum mean discrepancy (MMD) measure. 
\textsc{unioncom} performs unsupervised topological alignment for single-cell multi-omics data, emphasizing both local and global alignment. 
\textsc{scTopoGAN} uses topological autoencoders to obtain latent representations of each modality separately; a topology-guided Generative Adversarial Network then aligns these latent representations into a common space. We compare our method to \textsc{Infoot} \cite{chuang2023infoot} in \Cref{apds-snareseq}.
All alternative methods we consider require a choice of distance or similarity measures, which is a pain point that our method \ddmec~completely eliminates.

\label{sec:rna}

\paragraph{Datasets.}
We evaluate our method on single-cell multi-omics datasets: the peripheral blood mononuclear cells (\textsc{PBMC}) dataset and the bone marrow (\textsc{BM}) dataset. The \textsc{PBMC} dataset comprises healthy human peripheral blood mononuclear cells, profiled using the 10x Genomics multiome protocol, which enables simultaneous measurement of gene expression (RNA) and chromatin accessibility (ATAC) from the same cells. This dataset includes a total of 11{,}910 cells, encompassing 7 major immune cell types that are further subdivided into 20 finer-grained cell subclasses. The \textsc{BM} dataset consists of human bone marrow cells profiled using the CITE-seq protocol~\cite{stoeckius2017simultaneous}, which jointly captures gene expression (RNA) and protein abundance via antibody-derived tags (ADT). A set of 10{,}235 cells  are randomly selected  from each modality based on the major cell type labels.

For both datasets, we adopt the data preprocessing and evaluation pipeline described in~\cite{singh2023sctopogan}, resulting in 50-dimensional embeddings per modality. To assess the quality of the coupling, we compute cross-modal neighborhood consistency: for each cell in one modality, we identify its $k=5$ nearest neighbors in the aligned space from the other modality using Euclidean distance. We then evaluate the proportion of cases where the cell's class or subclass label matches the majority label among its neighbors. The resulting metrics are reported as the \emph{Celltype Acc} and \emph{Subcelltype Acc}, respectively. In this experiment, DDMEC is trained once, and inference is conducted five times with different
seeds.

\begin{table}[t]
\label{table_cell}
            \footnotesize
            \centering
            \begin{tabular}{lcc}
            \toprule
            \textbf{Method} & \emph{Celltype Acc} $\uparrow$ & \emph{Subcelltype Acc} $\uparrow$ \\ \midrule
                \multicolumn{3}{c}{\textsc{PBMC}} \\
            
            UnionCom$\star$ & 34.8 ± 10.9 & 22.9 ± 7.2 \\
             MMD-MA$\star$ & 28.3 ± 6.4 & 10.2 ± 4.8 \\
             SCOT $\star$& 12.9 ± 1.1 & 2.4 ± 0.2 \\
             scTopoGAN $\star$& 61.7 ± 8.6 & 41.3 ± 6.5 \\
             \textbf{\ddmec}  & \textbf{66.3 ± 2.6 } & \textbf{ 46.0 ± 0.5 } \\
              \midrule
            \multicolumn{3}{c}{\textsc{BM}} \\
           UnionCom $\star$& 51.8  ± 3.7 & 20.9 ± 2.6 \\
           MMD-MA$\star$ & 38.8 ± 17.9 & 10.4 ± 8.4 \\
           SCOT$\star$ & \textbf{90.5} ± 0.0 & 31.6 ± 0.0 \\
           scTopoGAN$\star$ & 50.9 ± 14.7 & 22.5  ± 5.4 \\
            \textbf{\ddmec} & \underline{77.3±0.1} & \textbf{44.2±0.1} \\
            \bottomrule
            \end{tabular}
            \vspace{0.5em}
            \caption{Single-Cell alignment experiments}
\end{table}

\paragraph{Results.} \Cref{table_cell} presents the quantitative results for the single-cell alignment task. Results marked with $\star$ are reported directly from~\citep{singh2023sctopogan}.  We observe that \ddmec~consistently outperforms existing baselines on the PBMC dataset, achieving superior performance in aligning both coarse-grained cell types and fine-grained cell subclasses. On the BM dataset, \ddmec~obtains the best performance for subclass-level alignment and ranks second for cell-type alignment.

\begin{figure}[h]
 \begin{subfigure}{0.45\textwidth}
        \centering
        \includegraphics[width=\linewidth]{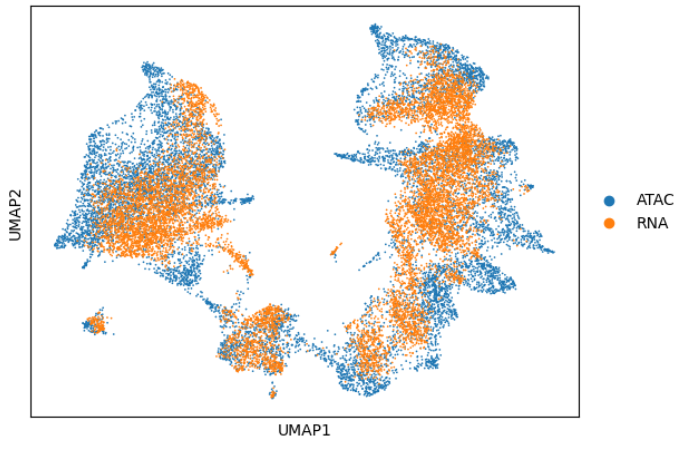}
    \end{subfigure}
    
    \begin{subfigure}{0.5\textwidth}
        \centering
        \includegraphics[width=\linewidth]{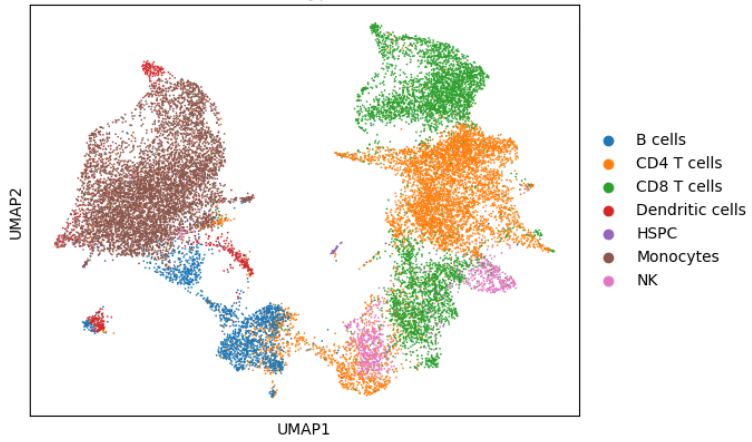}
    \end{subfigure}
    
\caption{Conditional generation with \ddmec~on \textsc{PBMC} dataset. \textbf{Top:} UMAP visualizations of  source and generated data. \textbf{Bottom:} points are colored by cell type, illustrating how well \ddmec~ preserves cell type separation.}
\label{fig:figure_pbmc}
\end{figure}

Notably, \ddmec~is the only method that demonstrates robust performance across both datasets, whereas alternative approaches exhibit inconsistent behavior—e.g., \textsc{scot} performs well on BM but fails to generalize to PBMC. In contrast to existing methods that learn deterministic one-to-one mappings between modalities, \ddmec~is fundamentally generative. It learns to sample from a coupling distribution rather than enforcing a fixed correspondence. To compute alignment metrics, we draw a sample in the target modality conditioned on a source cell, then identify the closest observed cell in the dataset using Euclidean distance. \Cref{fig:figure_pbmc} illustrates conditional generation with DDMEC using \textsc{UMAP} projections.

\subsection{Unpaired image translation}

This is a well-known problem in computer vision, where, in the absence of paired data (the joint distribution), the objective is to discover the correct mapping between two image domains. In this work, we show that unpaired image translation can be framed as a \gls{MEC} problem, where the goal is to learn the correct joint distribution between two unpaired image domains, $\mathcal{X}$ and $\mathcal{Y}$, respectively. Given the growing popularity of diffusion models in image-related tasks, pretrained weights for various image domains are available: we leverage them in our method \ddmec~, as done e.g. by~\citet{zhang2023adding}.

\paragraph{Baselines.} As the literature on image translation is vast, here we primarily focus on the unpaired case, and compare our method to a vast range of alternatives. \gls{GAN} have been widely applied to this domain~\citep{pang2021image}. These methods can be broadly categorized into those focusing on cycle-consistency, which enforces bidirectional mappings between image domains, such as \textsc{CycleGAN}~\citep{zhu2017unpaired}, \textsc{DualGAN}~\citep{yi2017dualgan}, \textsc{SCAN}~\citep{van2020scan}, and \textsc{U-GAT-IT}~\citep{kim2019u}; the second category uses distance-based methods, such as \textsc{DistanceGAN}~\citep{benaim2017one}, \textsc{GCGAN}~\citep{fu2019geometry}, \textsc{CUT}~\citep{park2020contrastive}, and \textsc{LSeSim}~\citep{zheng2021spatially}.
Diffusion-based models, which are related to our method, have also been explored for unpaired image translation. \textsc{UNIT-DDPM}~\citep{sasaki2021unit} learns two conditional models along with two additional domain translation models, incorporating a GAN-like cycle-consistency loss. \textsc{ILVR}~\citep{choi2021ilvr} and \textsc{SDEdit}~\citep{meng2021sdedit} utilize a diffusion model in the target domain while conditioning on a source image to refine the sampling procedure for image translation. \textsc{EGSDE}~\citep{zhao2022egsde} employs an energy function pretrained on both source and target domains to guide the inference process. Similarly, \textsc{SDDM}~\citep{sun2023sddm} introduces manifold constraints, forcing distributions at adjacent time steps to be decomposable into denoising and refinement components. Compared to these methods, \ddmec~leverages two conditional models, one per domain, which can be initialized using pretrained unconditional diffusion models. By design, our method does not require comparable domains and does not rely on a specific image similarity measure. We report results for two values of the guidance coefficient, a parameter influencing conditional generation.

\paragraph{Datasets.} We adopt the same experimental validation protocol as described by~\citet{zhao2022egsde}, where all images are resized to a resolution of 256 $\times$ 256. We use the \textsc{AFHQ}~\citep{stargan} dataset, consisting of high-resolution animal face images across three domains: \textsc{Cat}, \textsc{Dog}, and \textsc{Wild}. This dataset exhibits relatively large variations within and between domains, with 500 test images per domain. We compute the performance of our method \ddmec~and compare it to the baselines on \textsc{Cat}$\to$\textsc{Dog} and \textsc{Wild}$\to$\textsc{Dog} tasks. Furthermore, we employ the \textsc{CelebA-HQ}~\citep{gan_cleba} dataset, which comprises high-resolution human facial images categorized into two distinct domains: \textsc{Male} and \textsc{Female}. To evaluate the efficacy of our proposed approach relative to existing baselines, we conduct experiments on the domain translation task from \textsc{Male} to \textsc{Female}.

In addition, due to the bidirectional architecture of \ddmec - which leverages two conditional generative models - our framework also inherently supports translation in the reverse direction. The corresponding results for the \textsc{Female} $\to$ \textsc{Male} task are presented in~\Cref{app:additional_results}.

\paragraph{Results.} In~\Cref{image_res}, we present the quantitative results of \ddmec: results for alternative methods, marked with $\star$, are reported as obtained in~\citep{sun2023sddm,zhao2022egsde}. \ddmec~results are reported using 5 seeds.
The evaluation is based on generation quality, measured by the \gls{FID} score~\citep{heusel2017gans} (lower is better), and the fidelity to the source domain, assessed using \gls{SSIM} score~\citep{wang2004image} (higher is better). Note that quality and fidelity can be thought of as divergent objectives: high quality does not imply high fidelity and vice-versa.

On the \textsc{AFHQ} dataset, \gls{GAN}-based methods generally suffer from low image quality, except for \textsc{STARGAN}, which achieves a low \gls{FID} but performs poorly on \gls{SSIM}. In contrast, diffusion-based methods demonstrate superior performance compared to \gls{GAN}-based approaches. \ddmec~achieves the best \gls{FID} score in the \textsc{Cat}$\to$\textsc{Dog} task and the highest \gls{SSIM} in the \textsc{Wild}$\to$\textsc{Dog} task while maintaining comparable results on the remaining metrics. Overall, \ddmec~strikes the best balance between high-quality image generation and accurate alignment with the target domain.

Our  results on the \textsc{CelebA-HQ} dataset, demonstrate that \ddmec  outperforms competitors on both \gls{FID} and \gls{SSIM} even with only $50$ sampling steps. Specifically, at $50$ steps the \gls{FID} improves by approximately $1$ point and the \gls{SSIM} by $0.02$ points, with an even greater improvement (a $3$ point \gls{FID} reduction) when using $100$ sampling steps: it is well-known in the generative modeling literature that these improvements are significant. With the \textsc{CelebA-HQ} dataset, \ddmec benefits from a larger training set than in \textsc{AFHQ} animal dataset, and achieves state-of-the-art performance on image translation. 

\begin{figure}[h]
    \centering
 \begin{subfigure}{0.11\textwidth}
        \centering
        \includegraphics[width=\linewidth]{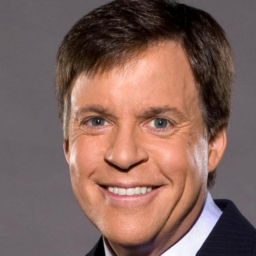}
    \end{subfigure}
    \begin{subfigure}{0.11\textwidth}
        \centering
        \includegraphics[width=\linewidth]{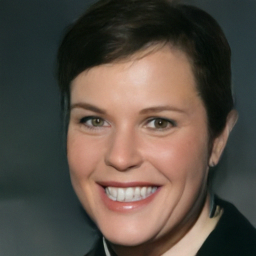}
    \end{subfigure} \hfill
     \begin{subfigure}{0.11\textwidth}
        \centering
        \includegraphics[width=\linewidth]{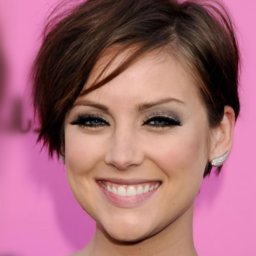}
    \end{subfigure} 
    \begin{subfigure}{0.11\textwidth}
        \centering
        \includegraphics[width=\linewidth]{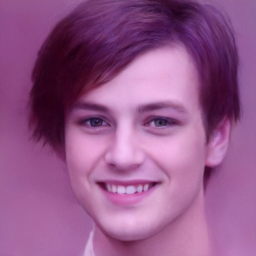}
    \end{subfigure}

\begin{subfigure}{0.11\textwidth}
        \centering
        \includegraphics[width=\linewidth]{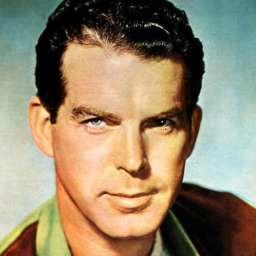}
    \end{subfigure}
    \begin{subfigure}{0.11\textwidth}
        \centering
        \includegraphics[width=\linewidth]{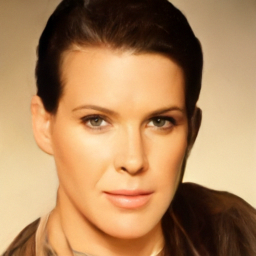}
    \end{subfigure} \hfill
    \begin{subfigure}{0.11\textwidth}
        \centering
        \includegraphics[width=\linewidth]{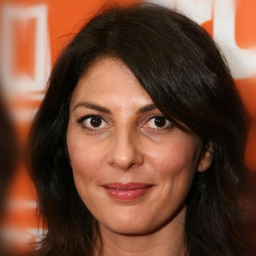}
    \end{subfigure} 
    \begin{subfigure}{0.11\textwidth}
        \centering
        \includegraphics[width=\linewidth]{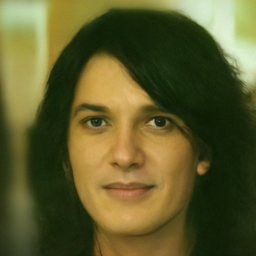}        
    \end{subfigure}

\begin{subfigure}{0.11\textwidth}
        \centering
        \includegraphics[width=\linewidth]{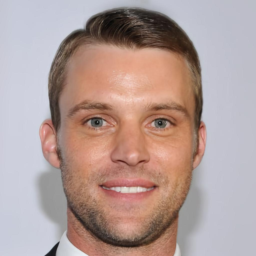}
    \end{subfigure}
    \begin{subfigure}{0.11\textwidth}
        \centering
        \includegraphics[width=\linewidth]{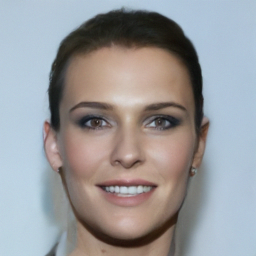}
    \end{subfigure} \hfill
    \begin{subfigure}{0.11\textwidth}
        \centering
        \includegraphics[width=\linewidth]{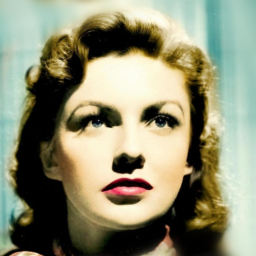}
    \end{subfigure} 
    \begin{subfigure}{0.11\textwidth}
        \centering
        \includegraphics[width=\linewidth]{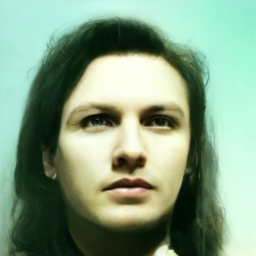}
    \end{subfigure}
\begin{subfigure}{0.11\textwidth}
        \centering
        \includegraphics[width=\linewidth]{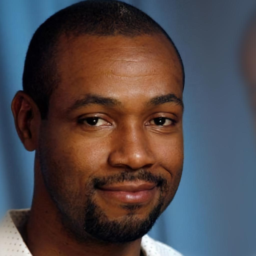}
        \caption*{Source}
    \end{subfigure}
    \begin{subfigure}{0.11\textwidth}
        \centering
        \includegraphics[width=\linewidth]{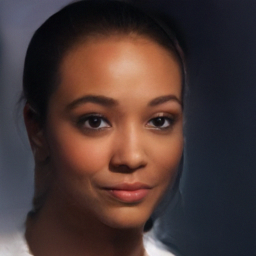}
        \caption*{Output}
    \end{subfigure} \hfill
    \begin{subfigure}{0.11\textwidth}
        \centering
        \includegraphics[width=\linewidth]{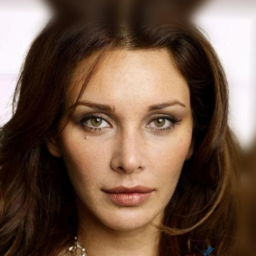}
         \caption*{Source}
    \end{subfigure} 
    \begin{subfigure}{0.11\textwidth}
        \centering
        \includegraphics[width=\linewidth]{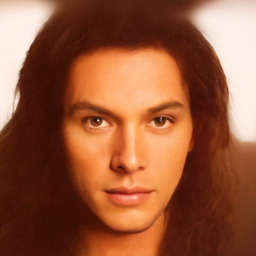}
        \caption*{Output}
    \end{subfigure}
\caption{Qualitative results of \ddmec-100~(guidance=2.5) on \textsc{CELEBA-HQ}. \textbf{Left:} :\textsc{Male}$\to$\textsc{Female} and \textbf{Right:} \textsc{Female}$\to$\textsc{Male} . Source domain image is used to generate the target female image.}
   \label{fig:celeba}

\end{figure}

\begin{table}[H]
\footnotesize
\caption{Quantitative image translation results.} 
\label{tab:quantitative comparison}
\begin{center}
 \begin{tabular}{lcc}
  \toprule
  Model & \acrshort{FID}$\downarrow$ & \gls{SSIM}$\uparrow$\\
  \midrule
   &\textsc{Cat}$\to$\textsc{Dog} & \\
  \midrule
  CycleGAN$\star$ &85.9 & -\\
  MUNIT$\star$ & 104.4 & -\\
  DRIT$\star$ & 123.4 & -\\
  Distance$\star$ & 155.3 & -\\
  SelfDistance$\star$ &144.4  & -\\
  GCGAN$\star$ & 96.6 & -\\
  LSeSim$\star$ &72.8  & -\\
  ITTR (CUT)$\star$ & 68.6 & -\\
  StarGAN v2$\star$ &\textbf{54.88 $\pm$ 1.01} & 0.27 ± 0.003 \\
  CUT$\star$ & 76.21 & \textbf{0.601} \\
  \midrule
  SDEdit$\star$ &74.17 $\pm$ 1.01 & \textbf{0.423 $\pm$ 0.001}\\
  ILVR$\star$ &74.37 $\pm$ 1.55 &0.363 $\pm$ 0.001 \\
  EGSDE$\star$ & 65.82 $\pm$ 0.77 & 0.415 $\pm$ 0.001\\
  SDDM$\star$ & 62.29 $\pm$ 0.63 & \textbf{0.422$\pm$ 0.001} \\
  \midrule
  50 Sampling steps & & \\
  \textbf{ \ddmec }(guidance=9)&  \textbf{60.70 $\pm$ 1.07} & 0.410 $\pm$ 0.001  \\
 \textbf{ \ddmec } (guidance=7) &  \textbf{58.50 $\pm$ 0.96} & 0.404 $\pm$ 0.001  \\
100 Sampling steps & & \\
\textbf{ \ddmec} (guidance=9) &  \textbf{60.51 $\pm$ 1.01} & 0.403 $\pm$ 0.001  \\
 \textbf{ \ddmec} (guidance=7) &  \textbf{57.89 $\pm$ 0.37} & 0.397 $\pm$ 0.001  \\

  \midrule
  &\textsc{Wild}$\to$\textsc{Dog} & \\
  \midrule
  SDEdit$\star$ & 68.51 $\pm$ 0.65 & 0.343 $\pm$ 0.001\\
  ILVR$\star$ & 75.33 $\pm$ 1.22 & 0.287 $\pm$ 0.001 \\
  EGSDE$\star$ & 59.75 $\pm$ 0.62 & 0.343 $\pm$ 0.001\\
  SDDM$\star$ & \textbf{57.38 $\pm$ 0.53}& 0.328 $\pm$ 0.001\\
   \midrule
  50 Sampling steps & & \\
  
  \textbf{ \ddmec }(guidance=9)&  62.03 $\pm$ 1.18 & \textbf{0.360 $\pm$ 0.002} \\
 \textbf{ \ddmec }(guidance=7)&   60.67 $\pm$ 1.01 & \textbf{0.353 $\pm$ 0.004}  \\

  100 Sampling steps & & \\
  \textbf{ \ddmec}(guidance=9)&  62.09 $\pm$ 0.59 & \textbf{0.356$\pm$  0.001} \\
 \textbf{ \ddmec}(guidance=7)&  59.22 $\pm$ 0.35 & \textbf{0.346 $\pm$ 0.001}  \\
  \midrule
  &\textsc{Male}$\to$\textsc{Female} & \\
  \midrule
  SDEdit$\star$ & 49.43 $\pm$ 0.47 & 0.572 $\pm$ 0.000 \\
  ILVR$\star$ & 46.12 $\pm$ 0.33 & 0.510 $\pm$ 0.001 \\
  EGSDE$\star$ & 41.93 $\pm$ 0.11 & 0.574 $\pm$ 0.000\\
  SDDM$\star$ & 44.37$\pm$ 0.23& 0.526 $\pm$ 0.001\\
   \midrule
  50 Sampling steps & & \\
\textbf{ \ddmec}(guidance=2.5)& \textbf{40.73}  $\pm$0.61  & \textbf{0.593 $\pm$ 0.003} \\
 \textbf{ \ddmec}(guidance=2)& \textbf{36.99}$\pm$ 0.83 & 0.556 $\pm$ 0.002 \\

  100 Sampling steps & & \\
 \textbf{ \ddmec}(guidance=2.5)&  \textbf{38.93  $\pm$ 0.37} & \textbf{0.588 $\pm$ 0.002}  \\
 \textbf{ \ddmec}(guidance=2)&  \textbf{34.86 $\pm$ 0.70} & 0.549 $\pm$ 0.002   \\
 \bottomrule
 \end{tabular}
\end{center}
\label{image_res}
\end{table}

\begin{figure}[h]
    \centering
    \begin{subfigure}[b]{0.11\textwidth}
        \centering
        \includegraphics[width=\textwidth]{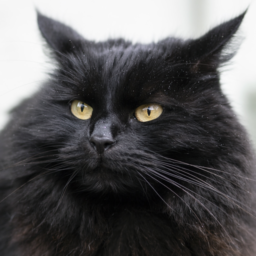}
    \end{subfigure}
    \begin{subfigure}[b]{0.11\textwidth}
        \centering
        \includegraphics[width=\textwidth]{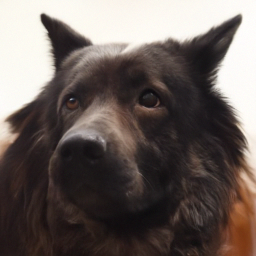}
    \end{subfigure} \hfill
    \begin{subfigure}[b]{0.11\textwidth}
        \centering
        \includegraphics[width=\textwidth]{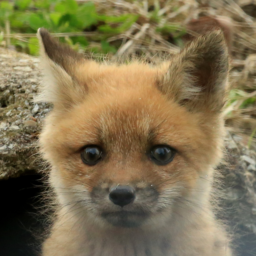}
    \end{subfigure}
    \begin{subfigure}[b]{0.11\textwidth}
        \centering
        \includegraphics[width=\textwidth]{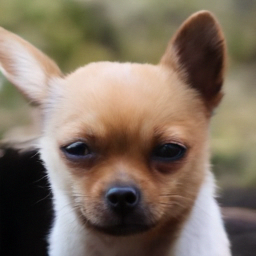}
    \end{subfigure}
    
     \begin{subfigure}[b]{0.11\textwidth}
        \centering
        \includegraphics[width=\textwidth]{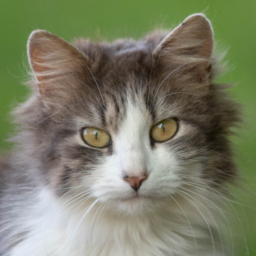}
  
    \end{subfigure}
    \begin{subfigure}[b]{0.11\textwidth}
        \centering
        \includegraphics[width=\textwidth]{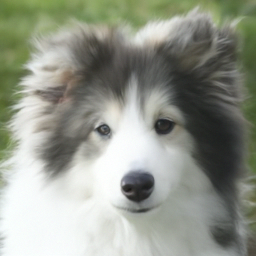}
    \end{subfigure}\hfill
    \begin{subfigure}[b]{0.11\textwidth}
        \centering
        \includegraphics[width=\textwidth]{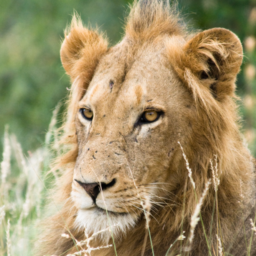}
    \end{subfigure}
    \begin{subfigure}[b]{0.11\textwidth}
        \centering
        \includegraphics[width=\textwidth]{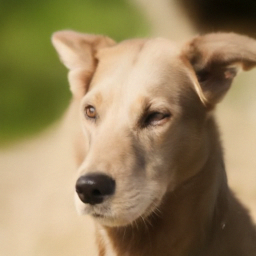}
    \end{subfigure}

    \begin{subfigure}[b]{0.11\textwidth}
        \centering
        \includegraphics[width=\textwidth]{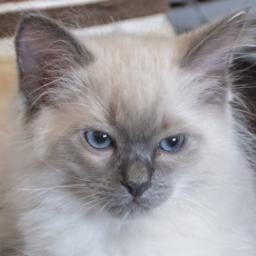}
    \end{subfigure}
    \begin{subfigure}[b]{0.11\textwidth}
        \centering
        \includegraphics[width=\textwidth]{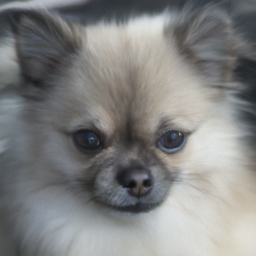}
    \end{subfigure}
    \hfill
    \begin{subfigure}[b]{0.11\textwidth}
        \centering
        \includegraphics[width=\textwidth]{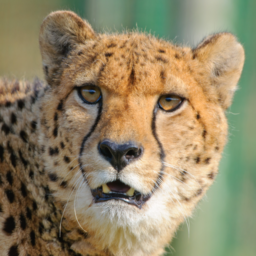}
    \end{subfigure}
    \begin{subfigure}[b]{0.11\textwidth}
        \centering
        \includegraphics[width=\textwidth]{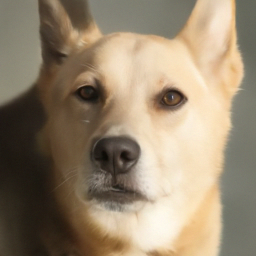}
    \end{subfigure}

\begin{subfigure}[b]{0.11\textwidth}
        \centering
        \includegraphics[width=\textwidth]{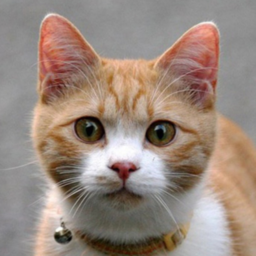}
    \end{subfigure}
    \begin{subfigure}[b]{0.11\textwidth}
        \centering
        \includegraphics[width=\textwidth]{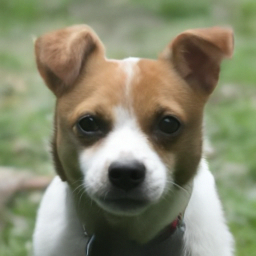}
    \end{subfigure}\hfill
 \begin{subfigure}[b]{0.11\textwidth}
        \centering
        \includegraphics[width=\textwidth]{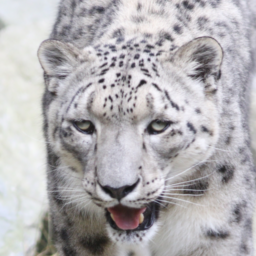}
    \end{subfigure}
    \begin{subfigure}[b]{0.11\textwidth}
        \centering
        \includegraphics[width=\textwidth]{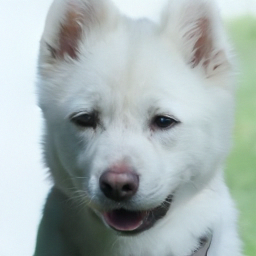}
    \end{subfigure}

\begin{subfigure}{0.11\textwidth}
        \centering
        \includegraphics[width=\linewidth]{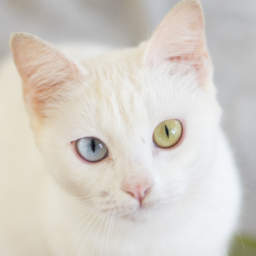}
    \end{subfigure}
    \begin{subfigure}{0.11\textwidth}
        \centering
        \includegraphics[width=\linewidth]{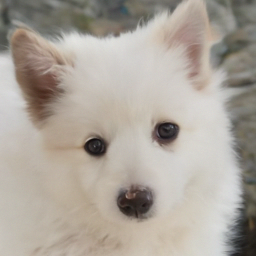}
    \end{subfigure}\hfill
    \begin{subfigure}{0.11\textwidth}
        \centering
        \includegraphics[width=\linewidth]{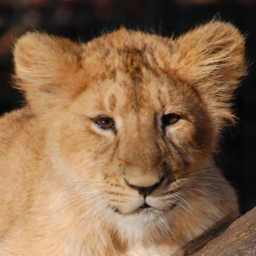}
    \end{subfigure} 
    \begin{subfigure}{0.11\textwidth}
        \centering
        \includegraphics[width=\linewidth]{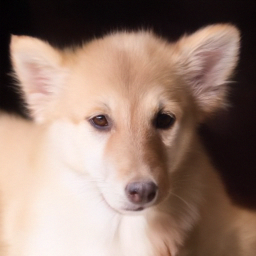}
    \end{subfigure}

\begin{subfigure}[b]{0.11\textwidth}
        \centering
        \includegraphics[width=\textwidth]{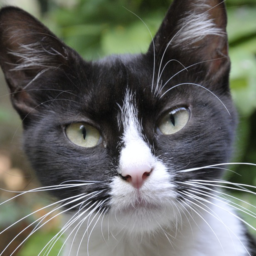}
         \caption*{Source}
    \end{subfigure}
    \begin{subfigure}[b]{0.11\textwidth}
        \centering
        \includegraphics[width=\textwidth]{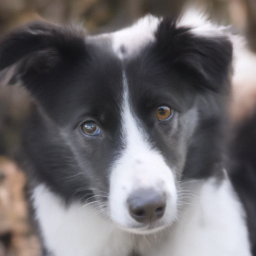}
           \caption*{Output}
    \end{subfigure}\hfill
 \begin{subfigure}[b]{0.11\textwidth}
        \centering
        \includegraphics[width=\textwidth]{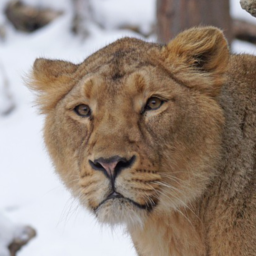}
        \caption*{Source}
    \end{subfigure}
    \begin{subfigure}[b]{0.11\textwidth}
        \centering
        \includegraphics[width=\textwidth]{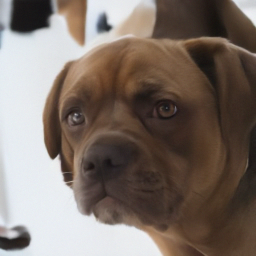}
        \caption*{Output}
    \end{subfigure}

    \caption{\ddmec~(guidance=7) \textsc{Cat}$\to$\textsc{Dog} (\textit{Left}) and \textsc{Wild}$\to$\textsc{Dog} image (\textit{right}) translation examples. Source domain image is used to generate the target dog image.}
  \label{fig:image_animals}
\end{figure}

This outcome aligns with expectations, as \ddmec~is designed to reduce uncertainty and enforce adherence to marginal constraints. In \cref{fig:image_animals,fig:celeba}, qualitative results further confirm the performance of \ddmec~in this task.
Additional results are available in \cref{fig:image_animals_add,fig:celeba_add}.

We investigate the effect of the guidance scale which acts as a temperature like parameter that controls conditioning strength—in our diffusion-based unpaired image translation framework. A test-time ablation study (\Cref{fig-guidance_ablation}) reveals a trade-off: higher guidance improves SSIM (structural similarity) but degrades FID (image realism), while lower guidance yields the opposite. This illustrates how guidance balances modality fidelity and mutual information maximization under the MEC framework.

\section{Conclusion}\label{sec:conclusion}
The machine learning community has recently directed substantial effort toward designing multimodal models, as they reflect the inherently multi-faceted nature of the real world. These models often achieve superior performance on downstream tasks compared to unimodal counterparts. However, the intrinsic complexity of multimodal data introduces significant challenges. In this work, we addressed the critical problem of coupling data represented by diverse modalities. The coupling problem has been widely studied in the literature, often framed as an optimal transport problem or approached with specialized architectures tailored to specific domains, such as images or language. However, existing methods typically rely on geometric spaces to compute costs, mappings, and similarities between data points.

We proposed a novel method that shifts the focus toward information and uncertainty quantification, thereby circumventing the limiting assumptions of prior approaches. Specifically, we studied the coupling problem through the lens of minimum entropy coupling. Since prior work on \gls{MEC} has largely been confined to discrete distributions, we extended this framework to continuous distributions.
Our key idea lies in introducing a parametric class of joint distributions reinterpreted as conditional generative models, augmented with terms to enforce adherence to marginal constraints. Our approach uses two models, which alternately optimize their objectives while approximately satisfying marginal constraints. 

The resulting method enables sampling and generation in either direction between modalities, without requiring specialized embeddings or strict geometric assumptions. 
Furthermore, it is adaptable to complex settings beyond one-to-one matching between modalities. 
We validated the performance of our approach in two domains. 
First, we applied it to multi-omics sequencing data, and we compared our method against several state-of-the-art alternatives that rely on predefined measures for data comparison and coupling cost definition. 
Our approach, being more general and free from stringent assumptions, achieves performance on par with or superior to these alternatives.
Second, we evaluated our method in the image translation domain, comparing it to a range of approaches from the literature. Our method demonstrated superior performance across widely recognized metrics for image quality and coherence, by striking a good balance between these often conflicting measures.


\clearpage{}
\section*{Acknowledgment} Pietro Michiardi was partially funded by project MUSE-COM$^2$ - AI-enabled MUltimodal SEmantic COMmunications and COMputing, in the Machine Learning-based Communication Systems, towards Wireless AI (WAI), Call 2022, ChistERA.

\section*{Impact Statement}
This paper aims to advance Machine Learning by improving unpaired data translation, with potential benefits for applications such as multi-omics alignment in biology and image translation. However, these advancements may also increase risks associated with generative models, including the generation of deepfakes. Responsible and ethical use is advised.
\bibliography{paper.bib}
\bibliographystyle{icml2025}

\clearpage{}
\appendix
\onecolumn
\section{Additional Details}\label{app:implementation_details}

\begin{figure}[h]
    \centering
    \includegraphics[width=1.0\textwidth]{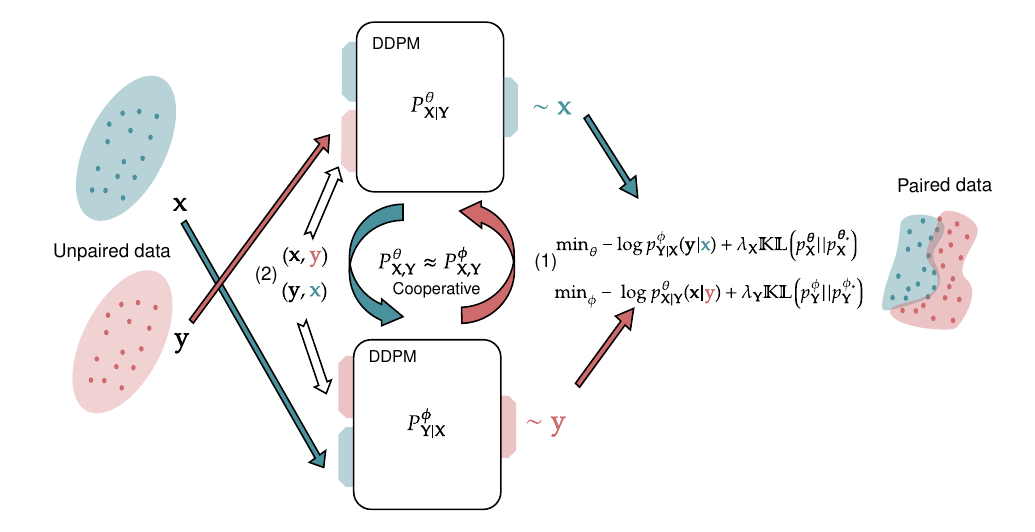}
    \label{ddmec_scheme}
     \caption[General scheme illustrating the \ddmec methodology]{Overview of the \ddmec methodology. The phase (1) corresponds to the procedure described in \Cref{algo:trainstep}, Lines~\ref{line2} and~\ref{line3}, and involves generating samples (depicted in red and blue) conditioned on inputs $x$ and $y$ drawn from their respective marginals. These samples are then used to evaluate the loss defined in \Cref{eq:DDMEC}, which is practically optimized using PPO \citep{dpok}. The phase (2) corresponds to Line~\ref{line4} of \Cref{algo:trainstep}, wherein the joint consistency constraint is enforced by updating the model with the previously generated sample pairs. Both phases require coordination between the two models, alternating their roles as outlined in \Cref{algo:trainloop}.}
     
\end{figure}

\subsection{Details on Swapping the Roles of the Conditional Models}
\label{apdx:proof}
In this part we provide more details about the justification of :
\begin{equation}\label{eqtoproove}
     \nabla_\theta \E_{x,y\sim p^\theta_{X,Y}}
[ - \log p^{\theta}_{X|Y}] \approx \nabla_\theta \E_{x,y\sim p^\theta_{X,Y}} [- \log p^{\phi}_{Y|Y}]
\end{equation}
\textit{whenever} $p^\theta_{X,Y} = p^\phi_{X,Y}$.

\Cref{eqtoproove} reads:
\begin{equation}
\nabla_\theta \int p_Y(y) \, p^\theta_{X|Y}(x|y) \log p^\theta_{X|Y}(x|y) \, dx \, dy.
\end{equation}

Moving the gradient \(\nabla_\theta\) inside the integral and applying the chain rule, we obtain:
\begin{align}
&\int p_Y(y) \, \nabla_\theta \left(p^\theta_{X|Y}(x|y)\right) \log p^\theta_{X|Y}(x|y) \, dx \, dy \nonumber \\
&\quad + \int p_Y(y) \, p^\theta_{X|Y}(x|y) \, \nabla_\theta \left(\log p^\theta_{X|Y}(x|y)\right) \, dx \, dy.
\end{align}

The second term simplifies to zero:
\begin{align}
\int p_Y(y) \, p^\theta_{X|Y}(x|y) \, \nabla_\theta \left(\log p^\theta_{X|Y}(x|y)\right) \, dx \, dy 
&= \int p_Y(y) \, \nabla_\theta \left(p^\theta_{X|Y}(x|y)\right) \, dx \, dy \nonumber \\
&= \nabla_\theta \int p_Y(y) \, p^\theta_{X|Y}(x|y) \, dx \, dy \nonumber \\
&= \nabla_\theta 1 = 0.
\end{align}

Assuming \( p^\theta_{X,Y} = p^\phi_{X,Y} \), \( p^\theta_X = p_X \), and \( p^\phi_Y = p_Y \), the first term rewrites as:
\begin{align}
&\int p_Y(y) \, \nabla_\theta \left(p^\theta_{X|Y}(x|y)\right) \log \frac{p^\phi_{Y|X}(y|x) p_X(x)}{p_Y(y)} \, dx \, dy \nonumber \\
&= \int p_Y(y) \, \nabla_\theta \left(p^\theta_{X|Y}(x|y)\right) \left( \log p^\phi_{Y|X}(y|x) + \log p_X(x) - \log p_Y(y) \right) \, dx \, dy \nonumber \\
&= \nabla_\theta \int p_Y(y) \, p^\theta_{X|Y}(x|y) \log p^\phi_{Y|X}(y|x) \, dx \, dy,
\end{align}
which corresponds to the right-hand side (r.h.s) of \Cref{eqtoproove}.

Indeed, the additional terms vanish:
\begin{align}
\int p_Y(y) \, \nabla_\theta \left(p^\theta_{X|Y}(x|y)\right) \log p_X(x) \, dx \, dy 
&= \int \nabla_\theta \left(p^\theta_X(x)\right) \log p_X(x) \, dx \nonumber \\
&= \int \nabla_\theta \left(p_X(x)\right) \log p_X(x) \, dx = 0,
\end{align}
and similarly,
\begin{equation}
-\int p_Y(y) \, \nabla_\theta \left(p^\theta_{X|Y}(x|y)\right) \log p_Y(y) \, dx \, dy = 0.
\end{equation}

\subsection{Diffusion Models Training with Reinforcement Learning}\label{app:dpok}

Our methodology begins by training unconditional diffusion models for both data modalities. We then use a reinforcement learning technique to train two conditional models (initialized from the first step) in a cooperative manner, allowing them to learn from each other to optimize the joint coupling under \gls{MEC} constraints and objectives. We formulate this second phase as training diffusion models with reinforcement learning and $\mathbb{KL}$-regularization. We follow the training scheme presented by~\citet{dpok}, where samples are generated conditionally using classifier guidance~\citep{ho2021classifierfree} with a DDIM sampler~\citep{song2020denoising}. The generated trajectories are then used to update the diffusion model, which is framed as a Markov Decision Process (MDP), using a policy gradient RL algorithm.

\paragraph{Reward Estimation}
In \ddmec, the reward signals are log-likelihood values mutually generated by the two conditional models. Accurately estimating this signal is crucial for steering training towards the optimal \gls{MEC} solution. To achieve this, we use multiple Monte Carlo steps to estimate Equation ~\ref{eq:likelihood}.

\paragraph{Policy Gradient Training}
We follow the training procedure of~\citet{dpok}, where, at each step, a batch of samples is generated using DDIM~\citep{song2020denoising}. These generated trajectories are then used to perform multiple gradient updates. Additionally, we apply importance sampling and ratio clipping~\citep{schulman2017proximal} to improve training stability.

\paragraph{Classifier-Free Guidance} We employ classifier-free guidance~\citep{ho2021classifierfree} in all experiments. This technique enables conditional sampling in step \ref{line3}. The denoising loss in \ref{line4} is optimized to account for the guidance mechanism by randomly dropping 10\% of the conditional signal, thereby stabilizing the unconditional model.

\subsection{Technical Details and Hyperparameters}
The source \footnote{\url{https://github.com/MustaphaBounoua/ddmec}} is publicly available.
\paragraph{Single-Cell Alignment}
For the \textsc{PBMC} and \textsc{BM} datasets, we utilize the preprocessed versions provided in the official code repository of \citet{singh2023sctopogan}\footnote{\url{https://github.com/AkashCiel/scTopoGAN}}, along with the accompanying evaluation protocols. Prior to training, we normalize the data by subtracting the mean and scaling to unit variance, while applying outlier mitigation. The model is first trained unconditionally using a \acrshort{DDPM} for 100{,}000 steps with $T = 1000$ diffusion steps. To stabilize training, we additionally incorporate a nearest neighbor retrieval step after each generation, where generated samples are projected back to the closest real data points using Euclidean distance. Subsequently, we train the two conditional models, following the procedure outlined in \Cref{algo:trainloop}. For each training step, we use a batch size of 256 and perform four gradient updates corresponding to line \ref{line3}, followed by four updates for Line \ref{line4}. We find it beneficial to accumulate generated samples during training and reuse them in optimizing Line \ref{line4}. We use a simple MLP network with skip connections and use the Adam optimizer~\citep{kingma2014adam} with a learning rate of $1 \times 10^{-4}$. The KL divergence regularization weight is set to $\lambda = 0.01$ for \textsc{PBMC} and $\lambda = 0.02$ for \textsc{BM}.



\paragraph{Unpaired Image Translation}

- \textbf{\textsc{Cat}$\to$\textsc{Dog} and \textsc{Wild}$\to$\textsc{Dog} Tasks:} We utilize the pre-trained model from the official implementation of~\citet{choi2021ilvr} (\url{https://github.com/jychoi118/ilvr_adm}) to initialize the dog modality conditional model.
For the other domains (\textsc{Cat}, \textsc{Wild}): We train a diffusion model from scratch using the same architecture and hyper-parameters as done in the target domain.
- \textbf{\textsc{Male}$\to$\textsc{Female} Task:} We use the publicly available pre-trained model by~\citet{zhao2022egsde} (\url{https://github.com/ML-GSAI/EGSDE}) for the Female modality. For the Male modality, we train a diffusion model from scratch using the same architecture and hyper-parameters as done in the target domain.

To introduce additional conditioning into the pre-trained diffusion model, we follow the work in~\citep{zhang2023adding}, where the encoder part of the \textsc{U-Net} is duplicated and used as a conditional encoder. The various hyperparameters are summarized in~\Cref{tab:hyperparams}. We follow the evaluation protocol described in~\citep{zhao2022egsde}.

\begin{table}[h]
\centering
\begin{tabular}{ccc} 
\hline
\toprule
\textbf{General Settings} &Dataset & \\
\midrule
& AFHQ  & CelebA-HQ \\
Batch Size & 16 &16 \\
Learning Rate: & $2e-5$ & $2e-5$\\
Optimizer & ADAM & ADAM\\
Training Steps & 2000 & \\
Weight Decay & 0.0  &\\
\hline
\textbf{Diffusion Model} &  \\
Noise Scheduler &  Linear &Linear \\
Number of Diffusion Timesteps ($T$) &  1000 & 1000\\
Sampler & DDIM & DDIM\\
Guidance Scale (training) &  7.0 & 7.0\\
Sampling steps &50 & 50\\
Exponential moving average & Yes & Yes\\
\hline
\textbf{Reinforcement Learning} &    \\
Reward (Monte Carlo steps) &3 & 3\\
Policy Update Steps&  4  & 4 \\
Importance Sampling Clipping & $1e-4$ &$1e-4$\\
$\lambda_1 , \lambda_2$ & $1e-3$ & $4e-3$ \\
Gradient Accumulation & 12 & 12 \\
Gradient Clipping & 1.0 &1.0\\
\hline
\end{tabular}
\caption{Hyperparameters used for training.}
\label{tab:hyperparams}
\end{table}
\newpage
\section{Additional Results}\label{app:additional_results}

\subsection{SNARE-seq additional experiments}
\label{apds-snareseq}

In this experiment we use the  \textsc{SnaReSeq}~\cite{chen2019high} dataset, which links chromatin accessibility with gene expression data on a mixture of four cells types.
We use the same preprocessing procedures detailed in~\cite{demetci2022scot}, which deal with filtering spurious data affected by technical errors, and normalization. 
Data samples have 1–1 correspondence information, which constitute the groud-truth information used for our performance evaluation. 
We use the average ``fraction of samples closer than the true match (\textsc{foscttm})'' metric introduced by~\citet{liu2019jointly}: given a sample in one domain, this amounts to compute the fraction of samples that are positioned more closely to it than its true match after pairing.
Results report the average \textsc{foscttm} across all samples, where lower values indicate better performance. 
We also report the label transfer accuracy as done by~\citet{cao2020unsupervised}, which measures how well sample labels are transferred between domains based on neighborhood alignment. A $k$-nearest neighbor classifier is trained on one domain and used to predict labels in the other. In this experiment, \ddmec~is trained once, and inference is conducted five times with different seeds. the \textsc{foscttm} metric, and is on-par with the best method in terms of accuracy. In this experiment, \ddmec~is trained once, and inference is conducted five times with different seeds. Unlike other methods, \ddmec~is conceptually different as it generates samples rather than learning a deterministic, 1-1 mapping. To compute the different metrics, given a sample from one modality, we use \ddmec~to generate a coupling to the other modality, then select the nearest sample from the dataset based on Euclidean distance.

\begin{table}[h]
\begin{center}{
\begin{tabular}{l *{4}{c}}
\toprule
 &  \multicolumn{2}{c}{\textbf{SNAREseq}}  \\
    & FOS $\downarrow$ & Acc $\uparrow$\\
    \midrule
    UnionCom$\star$  & 0.265 & 42.3
    \\
    MMD-MA$\star$  & 0.150 & 94.2
    \\
    SCOT$\star$  & 0.150 & 98.2
    \\
    InfoOT$\star$ & 0.156 & \textbf{98.8} \\
    \midrule
     \textbf{\ddmec} & \textbf{0.147}  & 98.6\\
    \bottomrule
      \end{tabular}}
\end{center}
\caption{Performance results on \textbf{SNAREseq} dataset}
\label{table_cell_snare}
\end{table}

\begin{figure}[ht]
    \centering
  \begin{subfigure}{0.4\textwidth}
        \centering
        \includegraphics[width=\linewidth]{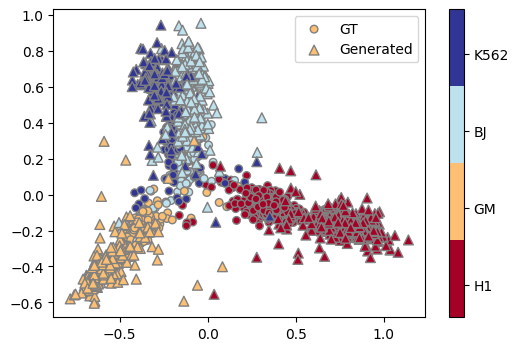}
    \end{subfigure}
    \begin{subfigure}{0.4\textwidth}
        \centering
        \includegraphics[width=\linewidth]{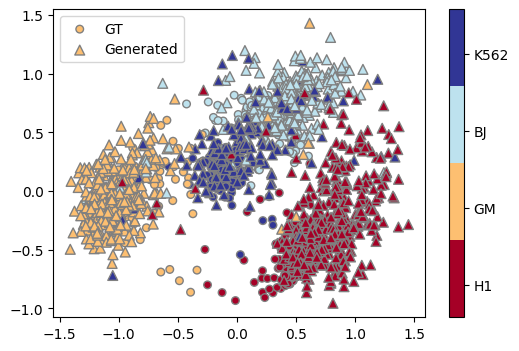}
    \end{subfigure}
\caption{Conditional generation using \ddmec~on the SNAREseq dataset. The cell types are indicated by colors. \textbf{Top:} generation of chromatin accessibility data using gene expression, \textbf{Bottom:} generation of gene expression using chromatin accessibility data.\label{fig:figure_snare}}
\end{figure}

\begin{figure}[ht]
    \centering
  \begin{subfigure}{0.15\textwidth}
        \centering
        \includegraphics[width=\linewidth]{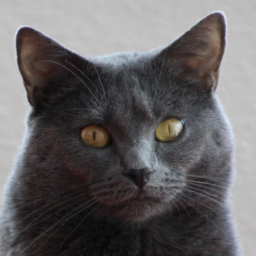}
    \end{subfigure}
    \begin{subfigure}{0.15\textwidth}
        \centering
        \includegraphics[width=\linewidth]{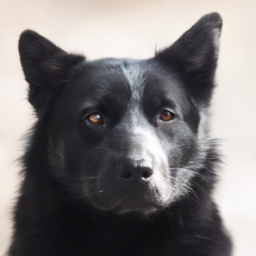}
    \end{subfigure}\hspace{2cm}
    \begin{subfigure}{0.15\textwidth}
        \centering
        \includegraphics[width=\linewidth]{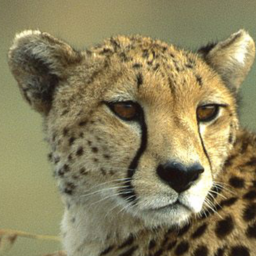}
    \end{subfigure} 
    \begin{subfigure}{0.15\textwidth}
        \centering
        \includegraphics[width=\linewidth]{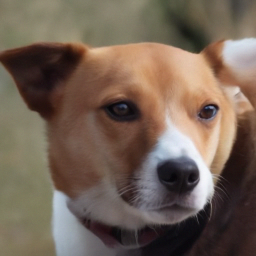}
    \end{subfigure}

\begin{subfigure}{0.15\textwidth}
        \centering
        \includegraphics[width=\linewidth]{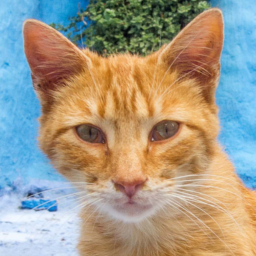}
    \end{subfigure}
    \begin{subfigure}{0.15\textwidth}
        \centering
        \includegraphics[width=\linewidth]{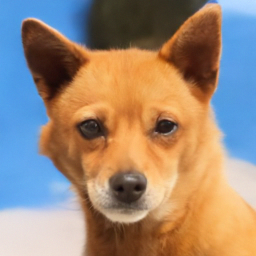}
    \end{subfigure}\hspace{2cm}
    \begin{subfigure}{0.15\textwidth}
        \centering
        \includegraphics[width=\linewidth]{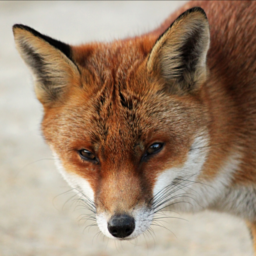}
    \end{subfigure} 
    \begin{subfigure}{0.15\textwidth}
        \centering
        \includegraphics[width=\linewidth]{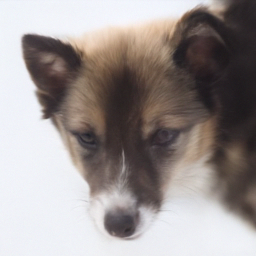}
    \end{subfigure}

\begin{subfigure}{0.15\textwidth}
        \centering
        \includegraphics[width=\linewidth]{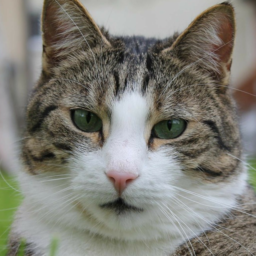}
    \end{subfigure}
    \begin{subfigure}{0.15\textwidth}
        \centering
        \includegraphics[width=\linewidth]{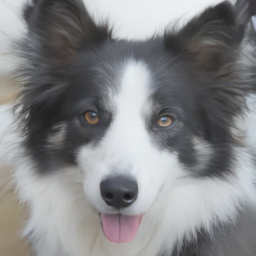}
    \end{subfigure}\hspace{2cm}
    \begin{subfigure}{0.15\textwidth}
        \centering
        \includegraphics[width=\linewidth]{assets/wild2dog/cond2/5.png}
    \end{subfigure} 
    \begin{subfigure}{0.15\textwidth}
        \centering
        \includegraphics[width=\linewidth]{assets/wild2dog/out2/5.png}
    \end{subfigure}

\begin{subfigure}{0.15\textwidth}
        \centering
        \includegraphics[width=\linewidth]{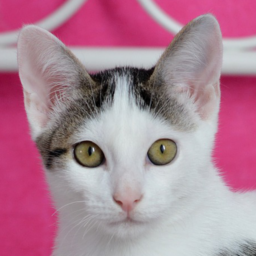}
    \end{subfigure}
    \begin{subfigure}{0.15\textwidth}
        \centering
        \includegraphics[width=\linewidth]{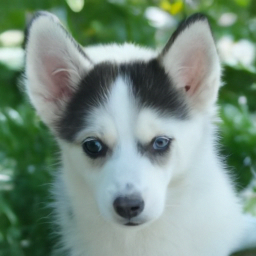}
    \end{subfigure}\hspace{2cm}
    \begin{subfigure}{0.15\textwidth}
        \centering
        \includegraphics[width=\linewidth]{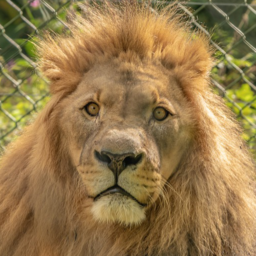}
    \end{subfigure} 
    \begin{subfigure}{0.15\textwidth}
        \centering
        \includegraphics[width=\linewidth]{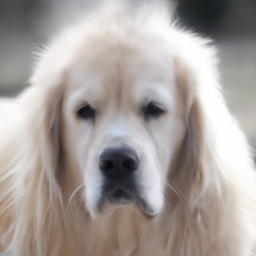}
    \end{subfigure}

\begin{subfigure}{0.15\textwidth}
        \centering
        \includegraphics[width=\linewidth]{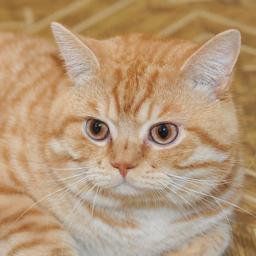}
    \end{subfigure}
    \begin{subfigure}{0.15\textwidth}
        \centering
        \includegraphics[width=\linewidth]{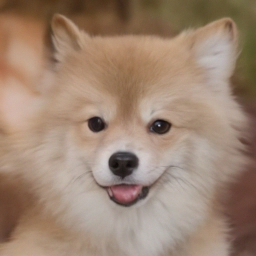}
    \end{subfigure}\hspace{2cm}
    \begin{subfigure}{0.15\textwidth}
        \centering
        \includegraphics[width=\linewidth]{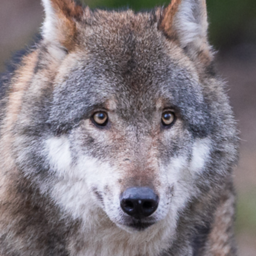}
    \end{subfigure} 
    \begin{subfigure}{0.15\textwidth}
        \centering
        \includegraphics[width=\linewidth]{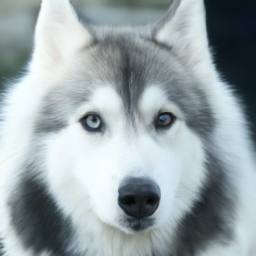}
    \end{subfigure}

\begin{subfigure}{0.15\textwidth}
        \centering
        \includegraphics[width=\linewidth]{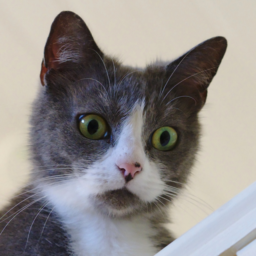}
    \end{subfigure}
    \begin{subfigure}{0.15\textwidth}
        \centering
        \includegraphics[width=\linewidth]{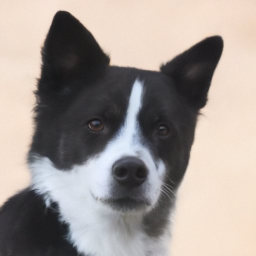}
    \end{subfigure}\hspace{2cm}
    \begin{subfigure}{0.15\textwidth}
        \centering
        \includegraphics[width=\linewidth]{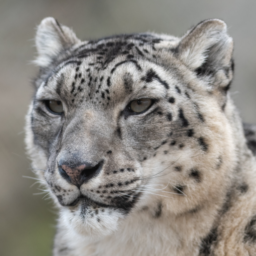}
    \end{subfigure} 
    \begin{subfigure}{0.15\textwidth}
        \centering
        \includegraphics[width=\linewidth]{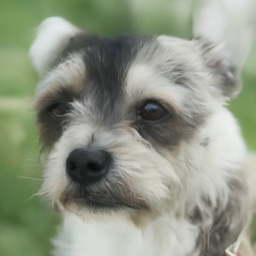}
    \end{subfigure}

    \begin{subfigure}{0.15\textwidth}
        \centering
        \includegraphics[width=\linewidth]{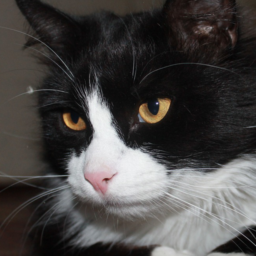}
    \end{subfigure}
    \begin{subfigure}{0.15\textwidth}
        \centering
        \includegraphics[width=\linewidth]{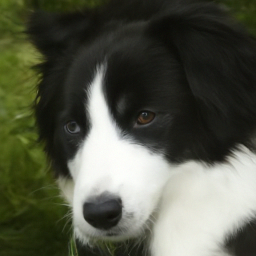}
    \end{subfigure}\hspace{2cm}
    \begin{subfigure}{0.15\textwidth}
        \centering
        \includegraphics[width=\linewidth]{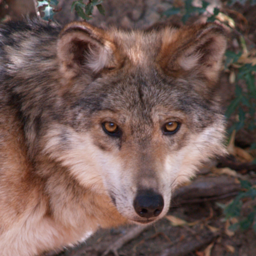}
    \end{subfigure} 
    \begin{subfigure}{0.15\textwidth}
        \centering
        \includegraphics[width=\linewidth]{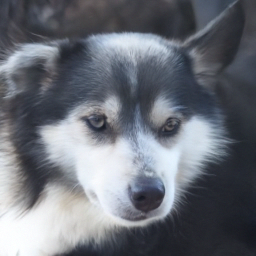}
    \end{subfigure}

    \begin{subfigure}{0.15\textwidth}
        \centering
        \includegraphics[width=\linewidth]{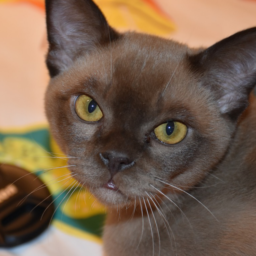}
            \caption*{Source}
    \end{subfigure}
    \begin{subfigure}{0.15\textwidth}
        \centering
        \includegraphics[width=\linewidth]{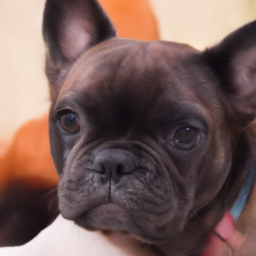}
            \caption*{Output}
    \end{subfigure}\hspace{2cm}
    \begin{subfigure}{0.15\textwidth}
        \centering
        \includegraphics[width=\linewidth]{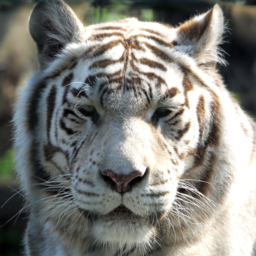}
            \caption*{Source}
    \end{subfigure} 
    \begin{subfigure}{0.15\textwidth}
        \centering
        \includegraphics[width=\linewidth]{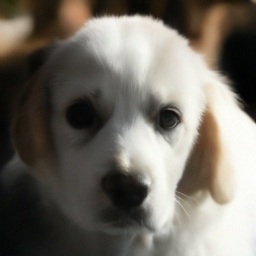}
            \caption*{Output}
    \end{subfigure}
     \caption{\ddmec~(guidance=7) \textsc{Cat}$\to$\textsc{Dog} (\textit{Left}) and \textsc{Wild}$\to$\textsc{Dog} image (\textit{right}) translation examples. Source domain image is used to generate the target dog image.}
  \label{fig:image_animals_add}
\end{figure}

\begin{figure}[ht]
    \centering
  \begin{subfigure}{0.15\textwidth}
        \centering
        \includegraphics[width=\linewidth]{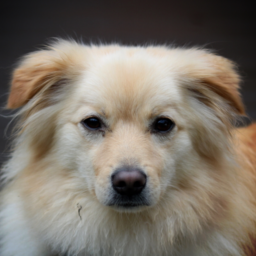}
    \end{subfigure}
    \begin{subfigure}{0.15\textwidth}
        \centering
        \includegraphics[width=\linewidth]{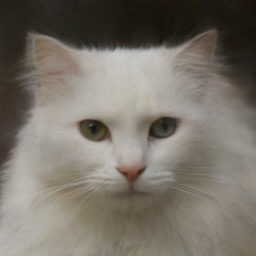}
    \end{subfigure}\hspace{2cm}
    \begin{subfigure}{0.15\textwidth}
        \centering
        \includegraphics[width=\linewidth]{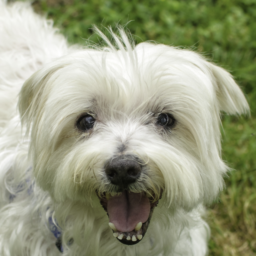}
    \end{subfigure} 
    \begin{subfigure}{0.15\textwidth}
        \centering
        \includegraphics[width=\linewidth]{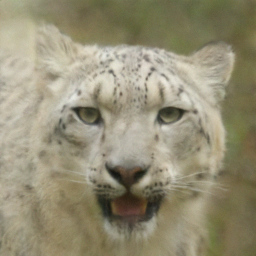}
    \end{subfigure}

    \begin{subfigure}[b]{0.15\textwidth}
        \centering
        \includegraphics[width=\textwidth]{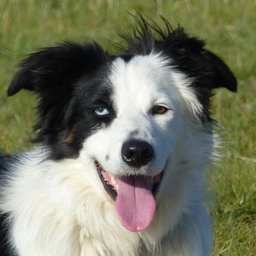}
    \end{subfigure}
    \begin{subfigure}[b]{0.15\textwidth}
        \centering
        \includegraphics[width=\textwidth]{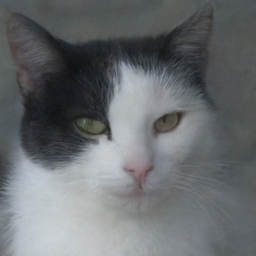}
    \end{subfigure} \hspace{2cm}
    \begin{subfigure}[b]{0.15\textwidth}
        \centering
        \includegraphics[width=\textwidth]{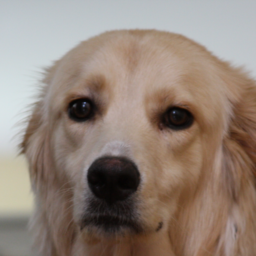}
    \end{subfigure}
    \begin{subfigure}[b]{0.15\textwidth}
        \centering
        \includegraphics[width=\textwidth]{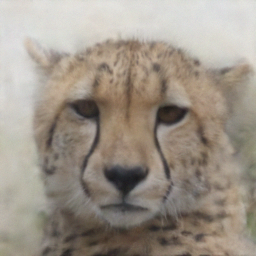}
    \end{subfigure}
    
     \begin{subfigure}[b]{0.15\textwidth}
        \centering
        \includegraphics[width=\textwidth]{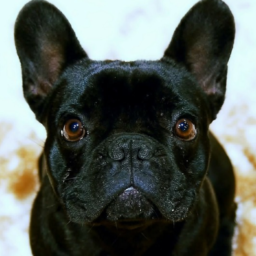}
  
    \end{subfigure}
    \begin{subfigure}[b]{0.15\textwidth}
        \centering
        \includegraphics[width=\textwidth]{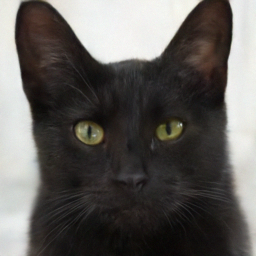}
    \end{subfigure}\hspace{2cm}
    \begin{subfigure}[b]{0.15\textwidth}
        \centering
        \includegraphics[width=\textwidth]{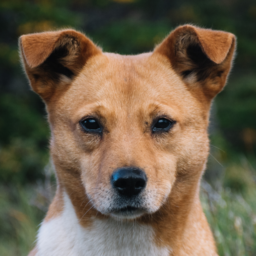}
    \end{subfigure}
    \begin{subfigure}[b]{0.15\textwidth}
        \centering
        \includegraphics[width=\textwidth]{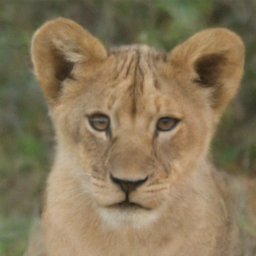}
    \end{subfigure}

    \begin{subfigure}[b]{0.15\textwidth}
        \centering
        \includegraphics[width=\textwidth]{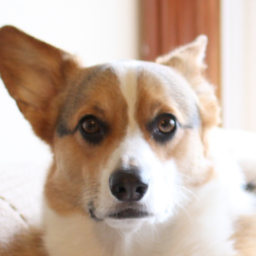}

    \end{subfigure}
    \begin{subfigure}[b]{0.15\textwidth}
        \centering
        \includegraphics[width=\textwidth]{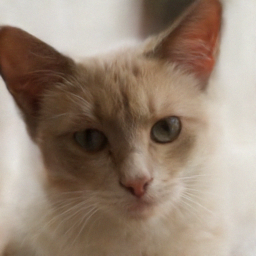}
    
    \end{subfigure}
    \hspace{2cm}
    \begin{subfigure}[b]{0.15\textwidth}
        \centering
        \includegraphics[width=\textwidth]{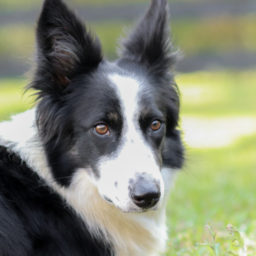}
    \end{subfigure}
    \begin{subfigure}[b]{0.15\textwidth}
        \centering
        \includegraphics[width=\textwidth]{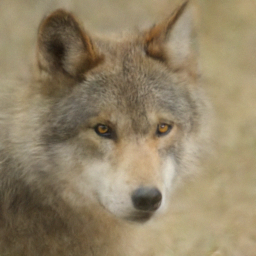}
    \end{subfigure}

\begin{subfigure}[b]{0.15\textwidth}
        \centering
        \includegraphics[width=\textwidth]{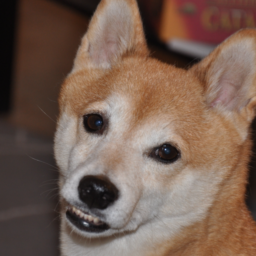}
    \end{subfigure}
    \begin{subfigure}[b]{0.15\textwidth}
        \centering
        \includegraphics[width=\textwidth]{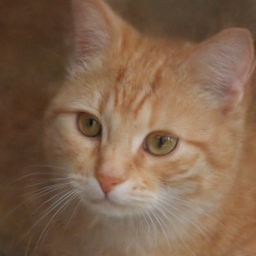}
    \end{subfigure}
    \hspace{2cm}
    \begin{subfigure}[b]{0.15\textwidth}\hfill
        \centering
        \includegraphics[width=\textwidth]{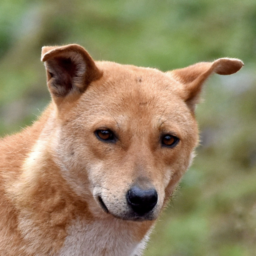}

    \end{subfigure}
    \begin{subfigure}[b]{0.15\textwidth}
        \centering
        \includegraphics[width=\textwidth]{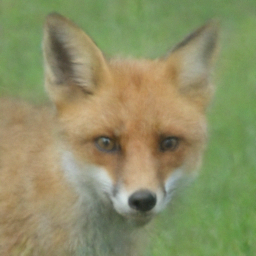}
    \end{subfigure}

    \begin{subfigure}[b]{0.15\textwidth}
        \centering
        \includegraphics[width=\textwidth]{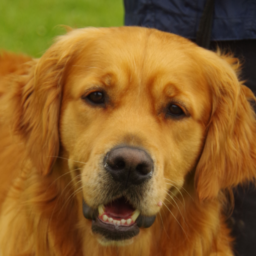}

    \end{subfigure}
    \begin{subfigure}[b]{0.15\textwidth}
        \centering
        \includegraphics[width=\textwidth]{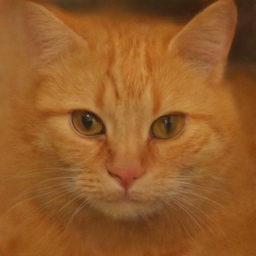}
    \end{subfigure}
    \hspace{2cm}
\begin{subfigure}[b]{0.15\textwidth}
        \centering
        \includegraphics[width=\textwidth]{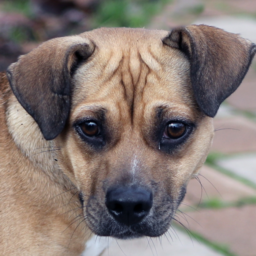}

    \end{subfigure}
    \begin{subfigure}[b]{0.15\textwidth}
        \centering
        \includegraphics[width=\textwidth]{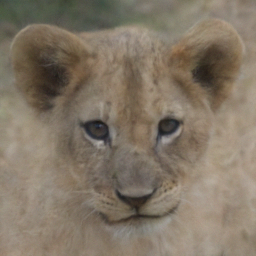}
    
    \end{subfigure}
     
\begin{subfigure}[b]{0.15\textwidth}
        \centering
        \includegraphics[width=\textwidth]{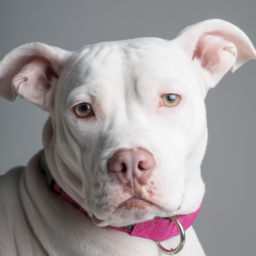}
    \end{subfigure}
    \begin{subfigure}[b]{0.15\textwidth}
        \centering
        \includegraphics[width=\textwidth]{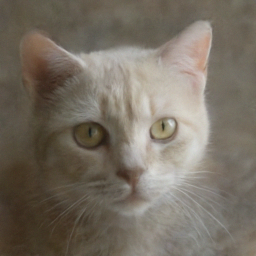}
    \end{subfigure}\hspace{2cm}
 \begin{subfigure}[b]{0.15\textwidth}
        \centering
        \includegraphics[width=\textwidth]{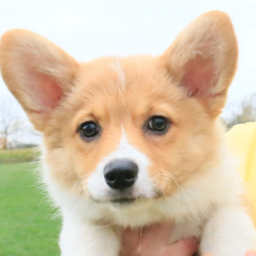}
    \end{subfigure}
    \begin{subfigure}[b]{0.15\textwidth}
        \centering
        \includegraphics[width=\textwidth]{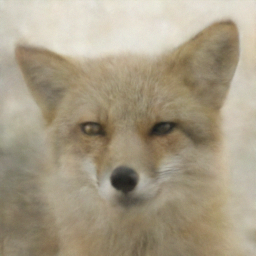}
    \end{subfigure}

     
\begin{subfigure}[b]{0.15\textwidth}
        \centering
        \includegraphics[width=\textwidth]{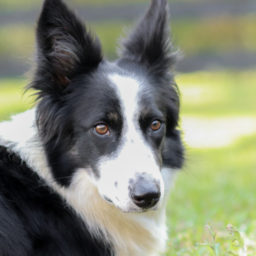}
              \caption*{Source}
    \end{subfigure}
    \begin{subfigure}[b]{0.15\textwidth}
        \centering
        \includegraphics[width=\textwidth]{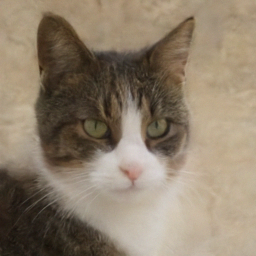}
              \caption*{Output}
    \end{subfigure}\hspace{2cm}
 \begin{subfigure}[b]{0.15\textwidth}
        \centering
        \includegraphics[width=\textwidth]{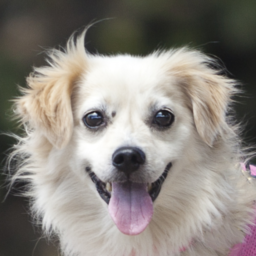}
            \caption*{Source}
    \end{subfigure}
    \begin{subfigure}[b]{0.15\textwidth}
        \centering
        \includegraphics[width=\textwidth]{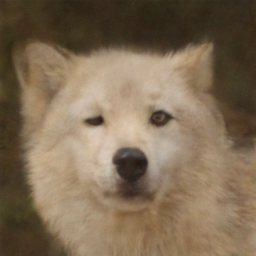}
            \caption*{Output}
    \end{subfigure}

    \caption{\ddmec~(guidance=7) \textsc{Dog}$\to$\textsc{Cat} (\textit{Left}) and \textsc{Dog}$\to$\textsc{Wild} image (\textit{right}) translation examples. Source domain image is used to generate the target Cat/Wild image.}
  \label{fig:image_animals_add2}
\end{figure}

\begin{figure}
\begin{subfigure}{0.7\textwidth}
     \centering
        \includegraphics[width=0.8\linewidth]{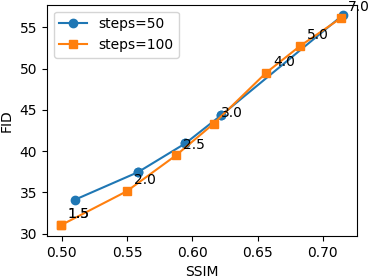}
        
  \caption{In the guidance scale ablation study, we report the FID and SSIM as a function of the guidance scale on the CelebA-HQ dataset. We notice that an increase in the guidance scale results in more information transfer between the two modalities, leading to a worse FID, and vice versa. }
  \label{fig-guidance_ablation}
\end{subfigure}
\end{figure}

\begin{figure}[ht]
    \centering

\begin{subfigure}{0.15\textwidth}
        \centering
        \includegraphics[width=\linewidth]{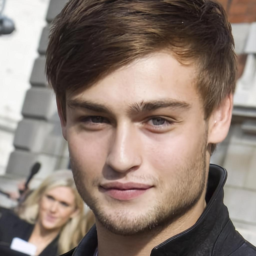}
    \end{subfigure}
    \begin{subfigure}{0.15\textwidth}
        \centering
        \includegraphics[width=\linewidth]{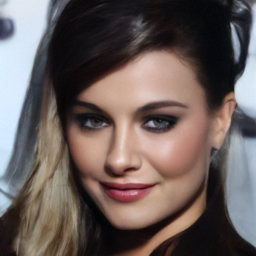}
    \end{subfigure} \hspace{2cm}
    \begin{subfigure}{0.15\textwidth}
        \centering
        \includegraphics[width=\linewidth]{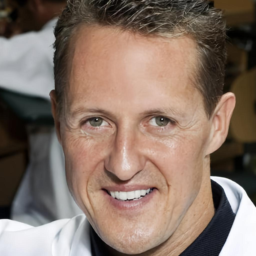}
    \end{subfigure} 
    \begin{subfigure}{0.15\textwidth}
        \centering
        \includegraphics[width=\linewidth]{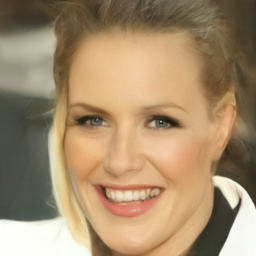}
    \end{subfigure}

\begin{subfigure}{0.15\textwidth}
        \centering
        \includegraphics[width=\linewidth]{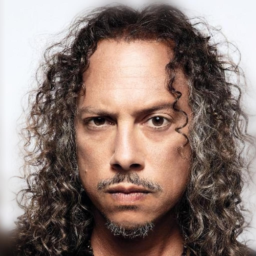}
    \end{subfigure}
    \begin{subfigure}{0.15\textwidth}
        \centering
        \includegraphics[width=\linewidth]{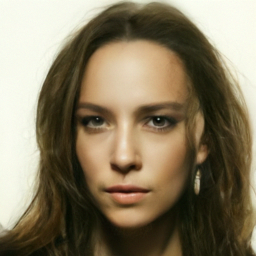}
    \end{subfigure} \hspace{2cm}
    \begin{subfigure}{0.15\textwidth}
        \centering
        \includegraphics[width=\linewidth]{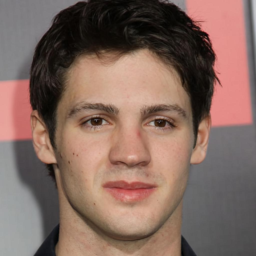}
    \end{subfigure} 
    \begin{subfigure}{0.15\textwidth}
        \centering
        \includegraphics[width=\linewidth]{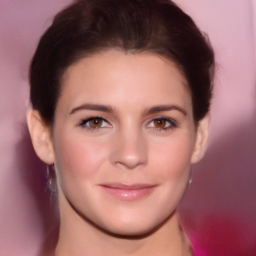}
    \end{subfigure}

\begin{subfigure}{0.15\textwidth}
        \centering
        \includegraphics[width=\linewidth]{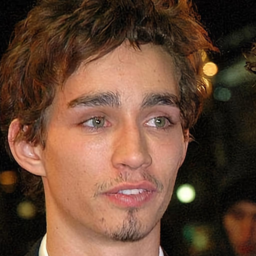}
    \end{subfigure}
    \begin{subfigure}{0.15\textwidth}
        \centering
        \includegraphics[width=\linewidth]{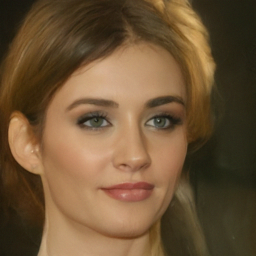}
    \end{subfigure} \hspace{2cm}
    \begin{subfigure}{0.15\textwidth}
        \centering
        \includegraphics[width=\linewidth]{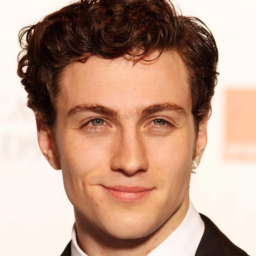}
    \end{subfigure} 
    \begin{subfigure}{0.15\textwidth}
        \centering
        \includegraphics[width=\linewidth]{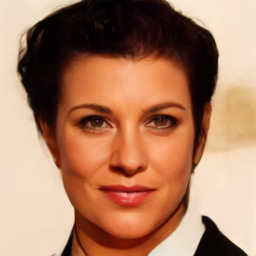}
    \end{subfigure}
    
\begin{subfigure}{0.15\textwidth}
        \centering
        \includegraphics[width=\linewidth]{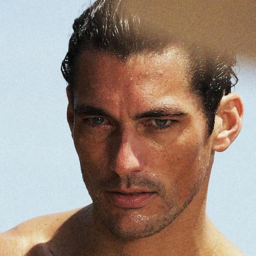}
    \end{subfigure}
    \begin{subfigure}{0.15\textwidth}
        \centering
        \includegraphics[width=\linewidth]{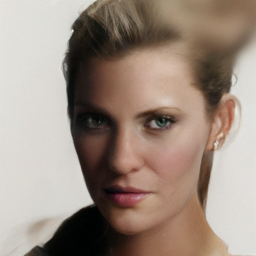}
    \end{subfigure} \hspace{2cm}
    \begin{subfigure}{0.15\textwidth}
        \centering
        \includegraphics[width=\linewidth]{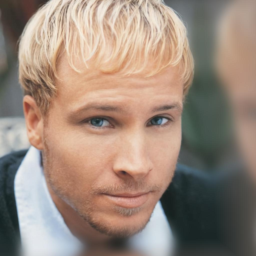}
    \end{subfigure} 
    \begin{subfigure}{0.15\textwidth}
        \centering
        \includegraphics[width=\linewidth]{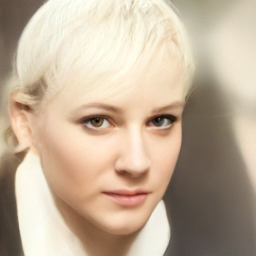}
    \end{subfigure}

  \begin{subfigure}{0.15\textwidth}
        \centering
        \includegraphics[width=\linewidth]{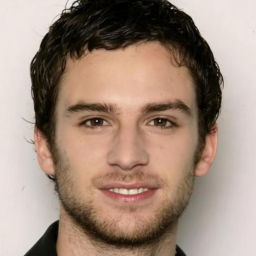}
                     \caption*{Source}
    \end{subfigure}
    \begin{subfigure}{0.15\textwidth}
        \centering
        \includegraphics[width=\linewidth]{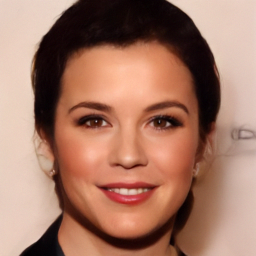}
             \caption*{Output}
    \end{subfigure}\hspace{2cm}
    \begin{subfigure}{0.15\textwidth}
        \centering
        \includegraphics[width=\linewidth]{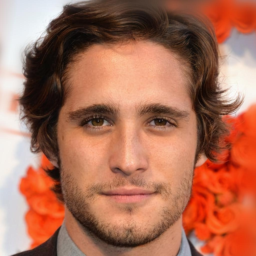}
        \caption*{Source}
    \end{subfigure} 
    \begin{subfigure}{0.15\textwidth}
        \centering
        \includegraphics[width=\linewidth]{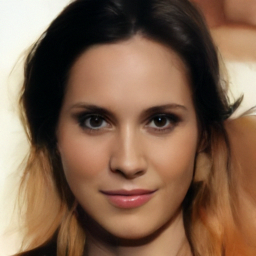}
             \caption*{Output}
    \end{subfigure}

           \caption{\ddmec~(guidance=2.5) \textsc{Male}$\to$\textsc{Female} translation examples. Source domain image is used to generate the target female image.}
   \label{fig:celeba_add}
\end{figure}

\end{document}